\documentclass{article}

\usepackage{arxiv}
\usepackage[utf8]{inputenc} 
\usepackage[T1]{fontenc}    
\usepackage{url}            
\usepackage{booktabs}       
\usepackage{amsfonts}       
\usepackage{xcolor}
\usepackage{graphicx}
\usepackage{amsmath}
\usepackage{upgreek}
\usepackage{subcaption}

\title{Statistical Agnostic Regression: a machine learning method to validate regression models}

\author{
  J.M. Gorriz, J. Ramirez, F. Segovia, F. J. Martinez-Murcia, C. Jiménez-Mesa and J. Suckling.\thanks{gorriz@ugr.es, jg825@cam.ac.uk} \\
  Data Science and Computational Intelligence Institute,  University of Granada, Spain \\
  Dpt. Psychiatry,  University of Cambridge, UK\\
 }

\begin{document}
\maketitle

\begin{abstract}
Regression analysis is a central topic in statistical modeling, aimed at estimating the relationships between a dependent variable, commonly referred to as the response variable, and one or more independent variables, i.e., explanatory variables. Linear regression is by far the most popular method for performing this task in various fields of research, such as data integration and predictive modeling when combining information from multiple sources. Classical methods for solving linear regression problems, such as Ordinary Least Squares (OLS), Ridge, or Lasso regressions, often form the foundation for more advanced machine learning (ML) techniques, which have been successfully applied, though without a formal definition of statistical significance. At most, permutation or analyses based on empirical measures (e.g., residuals or accuracy) have been conducted, leveraging the greater sensitivity of ML estimations for detection. In this paper, we introduce Statistical Agnostic Regression (SAR) for evaluating the statistical significance of ML-based linear regression models. This is achieved by analyzing concentration inequalities of the actual risk (expected loss) and considering the worst-case scenario. To this end, we define a threshold that ensures there is sufficient evidence, with a probability of at least $1-\eta$, to conclude the existence of a linear relationship in the population between the explanatory (feature) and the response (label) variables. Simulations demonstrate the ability of the proposed agnostic (non-parametric) test to provide an analysis of variance similar to the classical multivariate $F$-test for the slope parameter, without relying on the underlying assumptions of classical methods. Moreover, the residuals computed from this method represent a trade-off between those obtained from ML approaches and the classical OLS.
\end{abstract}


\section{Introduction}

Ordinary Least Squares (OLS) is the most popular method to perform linear regression analysis due to its optimal statistical properties assuming the linear model:
\begin{equation}\label{eq:uno}
    y=\upbeta_1^T \mathbf{x}+ \beta_0 +\epsilon 
\end{equation}
where $y$ is the response variable or observation, $\mathbf{x}$ is the $P\times 1$ explanatory variable or predictor, $\upbeta_{1}$ and $\beta_{0}$ are unknown parameters, slope and intercept, respectively,  that define the linear regressor (hyperplane) and $\epsilon$ is a random variable with zero mean and variance $\sigma^2$. The model above can be rewritten if we define $\upbeta=[\upbeta_1,^T \beta_0]^T$ and $\hat{\mathbf{x}}=[\mathbf{x}^T , 1]^T$ as:
\begin{equation}\label{eq:dos}
    y=\upbeta^T \hat{\mathbf{x}} +\epsilon 
\end{equation}
Using a set of observations $\mathbf{y}=[y_1,\ldots,y_N]^T$ and the predictor matrix $\mathbf{X}=[\hat{\mathbf{x}}_1,\ldots,\hat{\mathbf{x}}_N]^T$, the OLS estimator is obtained by minimizing the sum of squares $||\mathbf{y}-\mathbf{X}\upbeta||^2_2$. Following the Gauss-Markov theorem, the best linear unbiased estimator of any linear function is obtained by OLS if the assumptions of linearity, independence, homoscedasticity, and no perfect multicollinearity are fulfilled \cite{Zelen62}.

The link between OLS and machine learning (ML) regression methods lies in the fundamental principles of linear regression, which are shared by both traditional statistical approaches and more advanced ML techniques. ML regression methods, such as Ridge regression \cite{Hilt1977}, Lasso regression \cite{Tibshirani1996}, or Support Vector Machines (SVM) \cite{Burges98}, are formulated incorporating a regularizer $||\upbeta||^2$ into the minimization of the risk functional. In this context, the objective is to find a linear function $f(\hat{\mathbf{x}})=\upbeta^T\hat{\mathbf{x}}$ that minimizes the expected loss or actual risk $\mathcal{R}(f)=E[\mathcal{L}(f,\mathbf{x},y)]$, considering two terms. One term is akin to the OLS approximation and is referred to as the empirical risk, while the other is associated with model complexity. This insight is fundamental to the statistical learning theory (SLT), which serves as a strong foundation for all current artificial intelligence (AI) approaches \cite{Grohs22}. 

\subsection{Related Works}

Various approaches have been proposed to validate regression models \cite{Snee77}-\cite{Oredein11}.  \cite{Snee77} recommended comparing model predictions and coefficients with theoretical expectations, collecting new data, and employing techniques such as data splitting or cross-validation \cite{Kohavi95}. In \cite{Kleijnen96} the use of bootstrapping for validating metamodels was introduced, identified as a potent methodology whilst \cite{Miller91}, concentrated on logistic regression models and put forth a comprehensive approach that incorporated measures of goodness-of-fit, calibration, and refinement. Moreover, in \cite{Oredein11} a comparative analysis of the predictive index accuracy between data splitting and residual resampling bootstraps was conducted, concluding that bootstrapping yields a more precise estimate of goodness-of-fit indexes.

The main drawback of current AI approaches for classification and regression \cite{Lecun15,Gorriz2020} is their lack of rigorous analysis of significance in comparison to classical approaches such as the analysis of residuals in OLS linear regression using hypothesis testing. These approaches often limit their analyses to the use of permutation testing on empirical measures derived during the training stages in limited sample sizes, such as p-value analysis using cross-validation (CV). Moreover, although resampling techniques introduce little variation among bootstrap distributions, they can be quite variable, retaining the original variability of the random sample from the population. Randomization-based inference from a small sample may therefore be unreliable \cite{Moore2003}. For instance, numerous commentaries in the neuroimaging literature \cite{Varoquaux18,Eklund16,Jollans2019,Gorriz18,Braga04} highlight the high variability of performance across CV folds in various analytic designs, with clear implications for predictive inference. More critically, in pattern recognition, and particularly in regression problems, there is a significant concern about formulating ML analyses exclusively based on learning curves derived from selected loss functions \cite{Viering23} that merely demonstrate that the learning algorithm is converging to a potentially unreliable solution. 

Another explored possibility for testing significance is to apply classical analysis on the residuals produced by these AI approaches, which heavily relies on the assumptions of classical statistics, such as Gaussianity. Nonetheless, deep feature extraction analyses are typically conducted in low-dimensional spaces using linear classifiers \cite{vandermaaten08, Jimenez22, Gorriz2021}. In this paper, we propose the use of SLT to formulate a non-parametric statistical test for assessing the significance of ML regression models. Initially, we establish an upper bound on the actual risk (expected loss) of a (linear) support vector regressor under the worst-case scenario. Subsequently, we compare the bounded actual risk with that obtained under the null hypothesis $H_0$, similar to the 50\% rate in a classification problem, meaning there is no linear relationship between the predictors and the observed variables. Whenever the \emph{corrected} risk is less than this threshold, we reject $H_0$ and conclude that, with at least a probability $1-\eta$, there is a linear relationship between the predictor and observation.

\section{Background, definitions and notation:}
\label{sec:GLMLast}

The primary objective of SLT is to establish a structured approach for tackling the challenge of statistical inference, as highlighted in works such as \cite{Vapnik82,Haussler92}. We consider the supervised learning problem where a pair of predictors and observations $\{\mathbf{x}, y\}$ follows an unknown distribution $P$ that draws independent and identically distributed (iid) data. These pairs are used to estimate linear functions $f\in\mathcal{F}$ with a small risk.
\begin{equation}\label{eq:dos}
\mathcal{R}(f)=\int \mathcal{L}(f,\mathbf{x},y)dP(\mathbf{x},y),
\end{equation}
where $\mathcal{L}$ is  a loss function, e.g. $\mathcal{L}=(f(\mathbf{x})-y)^2$. The minimization of the actual risk is only possible through the evaluation of the empirical risk: 
\begin{equation}
    \mathcal{R}_N=E[\mathcal{L}(f,\mathbf{x},\mathbf{y})|(\mathbf{X},\mathbf{y})]
\end{equation}
given the sample $\mathbf{X}=\{\mathbf{x}_i\},\mathbf{y}=\{y_i\}$ for $i=1,\ldots,N$. 

A notable accomplishment of SLT is the development of straightforward and robust confidence intervals that effectively delimit the actual risk $\mathcal{R}(f)$ \cite{Boucheron13}. Our specific focus lies in the estimation of risk derived from an empirical quantity, ensuring a probability of at least $1-\eta$. This estimation is articulated through the concentration inequalities (CI) presented as:
\begin{equation}\label{eq:dos}
\mathcal{R}(f_N)\leq\mathcal{R}_N(f_N)+\Delta(N,\mathcal{F})
\end{equation}
where $f_N$ is carefully chosen to prevent overfitting by restricting the class of functions $\mathcal{F}$, such that:\begin{equation}
    f_N=\underset{f\in\mathcal{F}}{ \arg\min }\mathcal{R}_N(f)
\end{equation} 
and the term $\Delta(N,\mathcal{F})$ acts as an upper bound for the actual risk, depending on the complexity of the class $\mathcal{F}$ and the sample $Z_N=\{\mathbf{X},\mathbf{y}\}$.  In the worst-case scenario, this inequality becomes an equality. This deviation can be understood through various perspectives in classical probability theory offering insights into how closely the sum of independent random variables (empirical risk) are to their expectations (actual risk) \cite{Vapnik82}.

\subsection{Support vector regression: theory and practice}\label{sec:background}

The Support Vector Regression (SVR) algorithm can be generalized to the case of regression estimation \cite{Vapnik95},  where the sparseness property is preserved by the definition of the $\epsilon$-insensitive loss function:
\begin{equation}\label{eq:tres}
    |y-f(\mathbf{x})|_\epsilon=\max\{0,|y-f(\mathbf{x})|-\epsilon\},
\end{equation}
where the parameter $\epsilon$ is automatically computed \cite{Scholkopf95}. The expected loss to be minimized is the regularized risk functional:
\begin{equation}\label{eq:cuatro}
   \frac{1}{2}||\upbeta||^2+\frac{c}{N}\sum_{i=1}^N |y_i-f(\mathbf{x_i})|_\epsilon
\end{equation}
Observe that $c$ is the trade-off constant between complexity and training error. The minimization of equation \ref{eq:cuatro} is equivalent to a constrained optimization problem detailed elsewhere \cite{Burges98}.  It is important to note that the choice of loss function depends on the specific regression problem at hand \cite{Huber64}. In this paper, we will employ the two most commonly used losses: the squared loss (also known as $\mathcal{L}_2$) and the absolute value loss or $\mathcal{L}_1$ (the insensitive loss for $\epsilon$=0). However, previous studies have indicated that when additive noise deviates from Gaussianity, better approximations to the regression problem are achieved by using estimators based on alternative loss functions, beyond the quadratic loss function employed in OLS \cite{Huber64, Vapnik95}. To make the two losses comparable, we will employ the rounded quadratic loss in the case of the OLS algorithm whilst ML models are using the least modulo method, e.g., $\epsilon$-insensitive loss. Finally,  we point out that the quadratic problem associated with equation  \ref{eq:cuatro} is connected to support vector classification \cite{Scholkopf95}.  Furthermore, the hyperplane that is constructed by this minimization, and lies close to as many of the data points as possible,  is a self-supervised training strategy to classify the samples outside the $\epsilon$-tube. 

\subsubsection{Flatness and the Empirical Loss in Testing for Linearity}

\begin{figure*}
\centering
\begin{subfigure}{0.49\textwidth}
    \includegraphics[width=\textwidth]{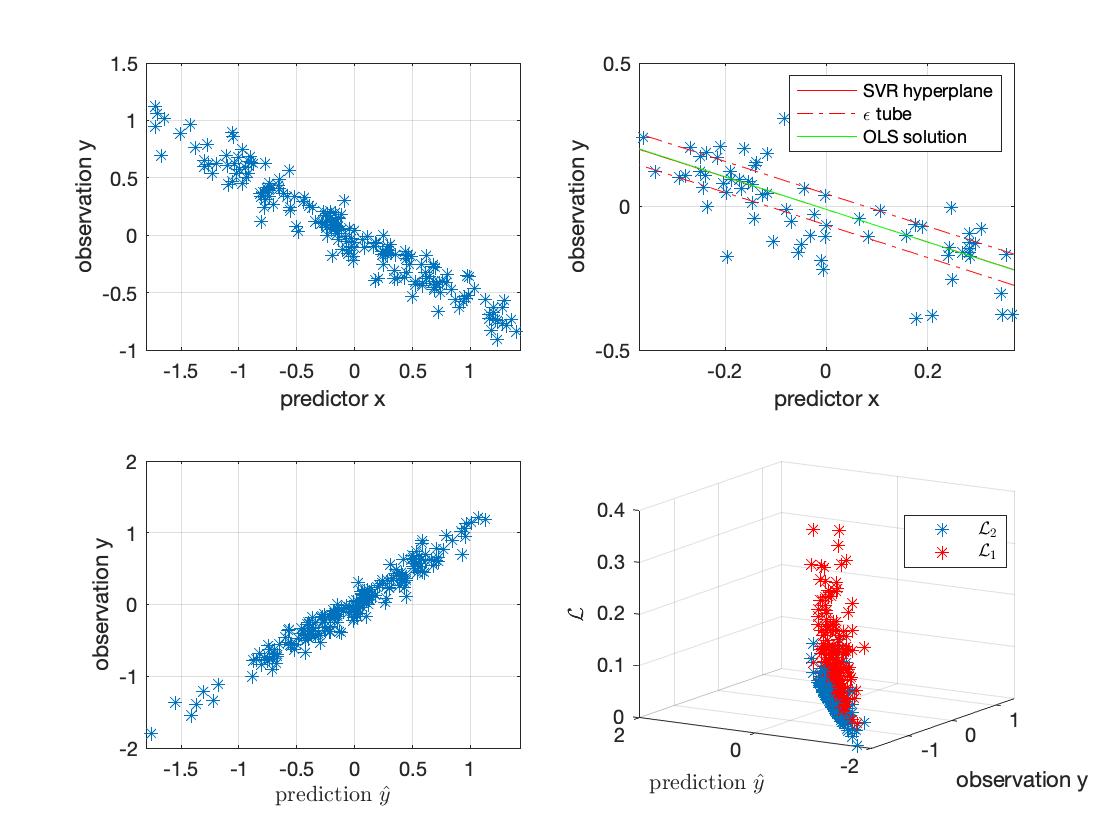}
    \caption{Regression fitting for $\tau=0.9$.}
    \label{fig:1a}
\end{subfigure}
\hfill
\begin{subfigure}{0.49\textwidth}
    \includegraphics[width=\textwidth]{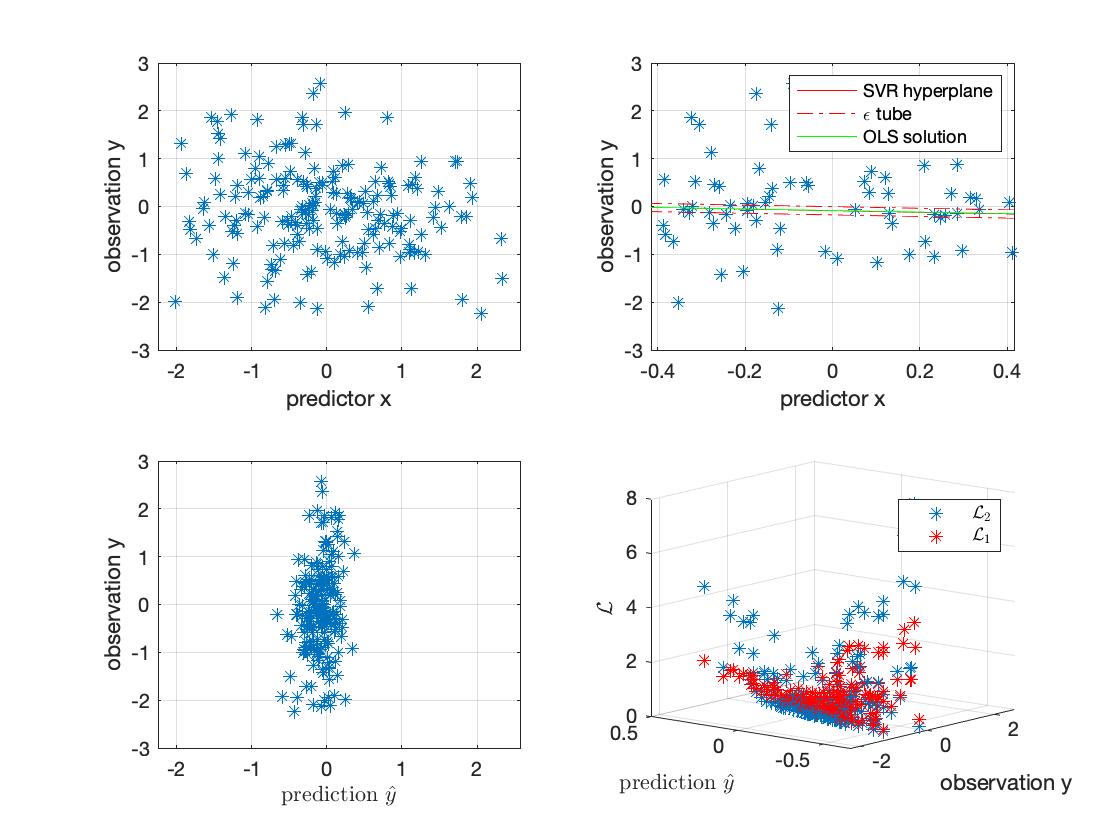}
    \caption{Regression fitting for $\tau=0.1$.}
    \label{fig:1b}
\end{subfigure}
\hfill
\begin{subfigure}{0.32\textwidth}
    \includegraphics[width=\textwidth]{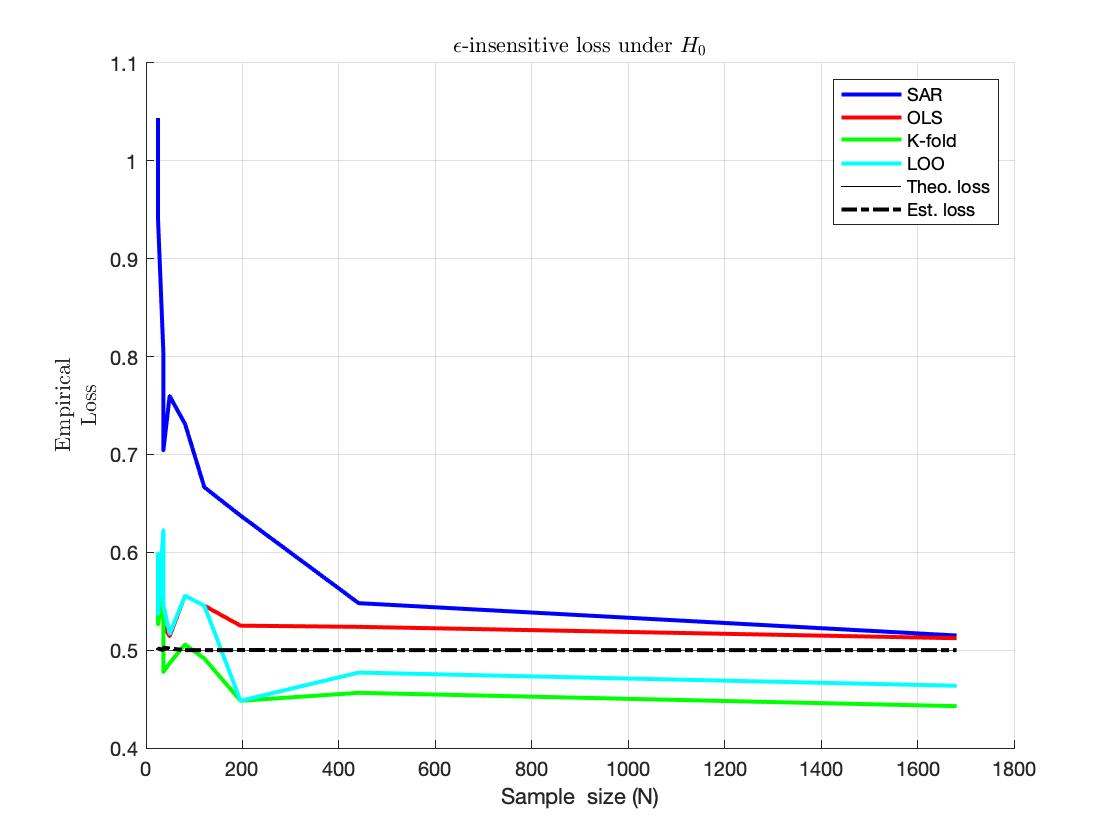}
    \caption{Averaged (empirical) loss under $H_0$ ($\mathcal{L}_1$).}
    \label{fig:1c}
\end{subfigure}
\hfill
\begin{subfigure}{0.32\textwidth}
    \includegraphics[width=\textwidth]{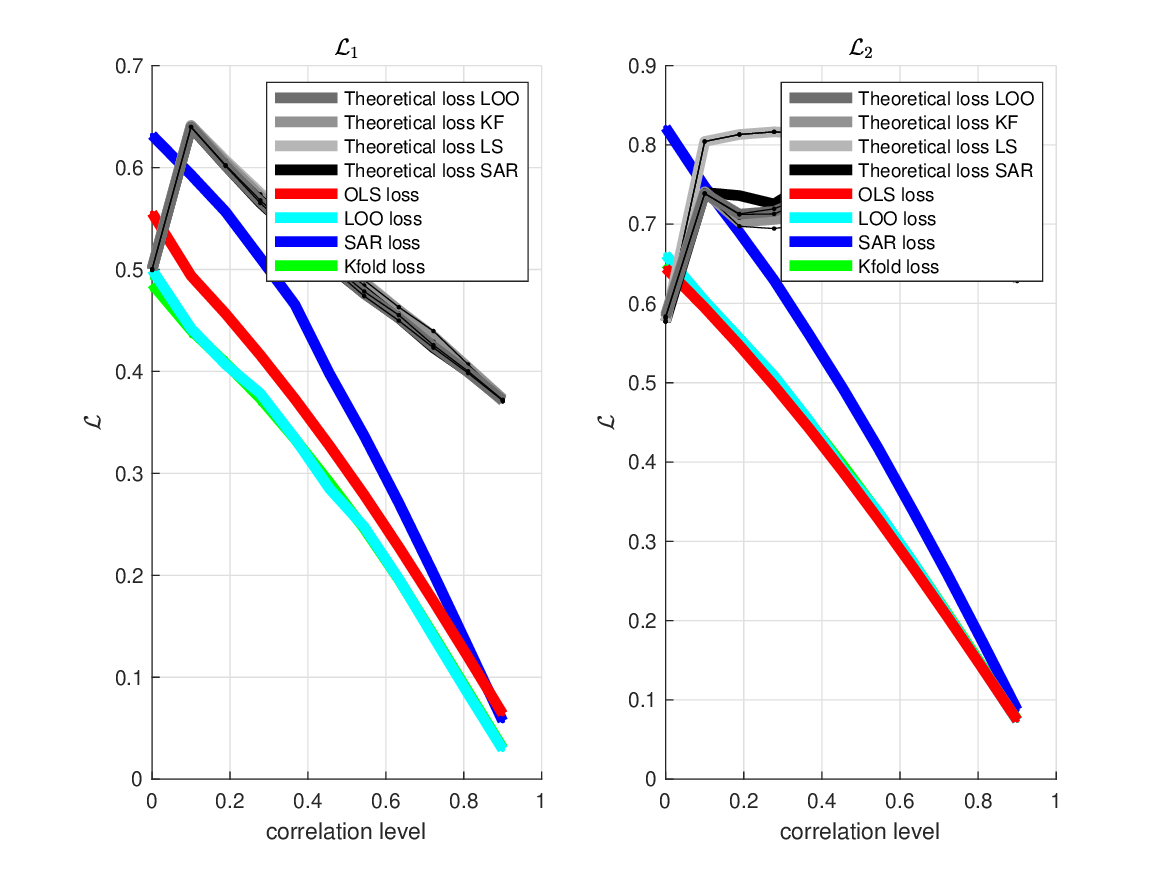}
    \caption{Averaged losses vs correlation for $\mathcal{L}_{1,2}$}
    \label{fig:1d}
\end{subfigure}
\hfill
\begin{subfigure}{0.32\textwidth}
    \includegraphics[width=\textwidth]{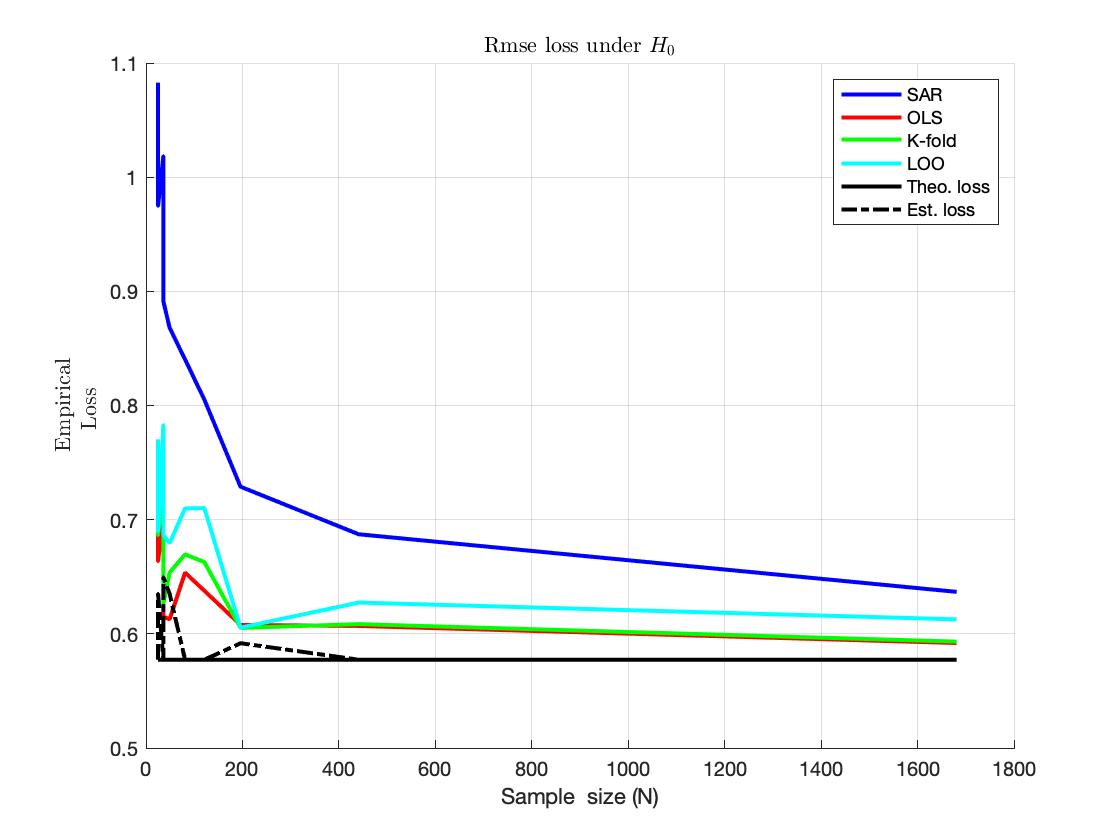}
    \caption{Averaged (empirical) loss under $H_0$ ($\mathcal{L}_2$)..}
    \label{fig:1e}
\end{subfigure}
        
\caption{In this 2D example of regression fitting, we explore effects ranging from large to very small. Observe the flatness of the linear functions in subfigure \ref{fig:1b}. In the regression examples of subfigures \ref{fig:1a} and \ref{fig:1b}, we present several representations: y vs. x; the result of regression; y vs. $\hat{y}$; and $\mathcal{L}$ vs. (y, $\hat{y}$). In figures \ref{fig:1c} and \ref{fig:1e}, we plot the empirical losses for all the methods using uncorrelated data within a mesh grid and compare them with the theoretical value. In the middle figure, we demonstrate how the theoretical losses under $H_0$ for all the tested methods envelop the empirical losses, except for the case of uncorrelated data where the correlation level is equal to zero.}.
\label{fig:example1}
\end{figure*}

As an example, and connected to the novel test proposed in the next section, let's assume we are provided with a dataset of observations $y$ and predictors $x$ in 2 dimensions. Starting from an i.i.d. 2D Gaussian distribution, we can define scaling and rotation transforms $\mathbf{T} = \mathbf{S}\mathbf{R}$, where $\mathbf{S} = \text{diag}(1, 1 - \tau)$,  where $\tau \in (0,1)$ is the correlation level, and $\mathbf{R} =  \begin{pmatrix}  \cos(\theta) & -\sin(\theta) \\
 sin(\theta) & \cos(\theta) \end{pmatrix}$.  These transforms are applied to the data to create a non-diagonal covariance distribution, i.e.  a linear relationship (see figures \ref{fig:1a} and \ref{fig:1b}).Then, we assess the individual losses and the expected loss (average) of the problem in the same figures. 

From these examples, we readily see that with decreasing correlation level, the flatness property of the OLS and SVR solutions dominate over the minimization of the empirical risk that, with uncorrelated data, provides an extreme value. For instance, within the framework of the OLS method, the expression for the squared loss function, denoted as $\mathcal{L}_2$, is given by $(f(\mathbf{x}) - y)^2$. Recall that $\boldsymbol{\beta}$ is determined by the OLS solution $\boldsymbol{\beta} = (\mathbf{X}^T\mathbf{X})^{-1}\mathbf{X}^T\mathbf{y}$ when the rank of the design matrix $\mathbf{X}$ is equal to the number of observations $N$ \cite{Zelen62}. The solution becomes null ($\boldsymbol{\beta} = \mathbf{0}$) when $\mathbf{X}^T\mathbf{y} = \mathbf{0}$, indicating a perfect orthogonality issue in the dataset, which is the opposite of multicollinearity.  Therefore, the expected loss simplifies to $\mathcal{R}_N|_{\upbeta=0} = \frac{1}{N}(\mathbf{y} - \mathbf{X}\upbeta)^T(\mathbf{y} - \mathbf{X}\upbeta) = \frac{1}{N}\mathbf{y}^T\mathbf{y}$.  

In this hypothetical case, the effect of sampling and sample size is depicted in figures \ref{fig:1c}-\ref{fig:1e}, where we additionally illustrate individual and averaged losses converging to the theoretical value. This theoretical value (gray curves in figure \ref{fig:1d}) could serve as a threshold to establish the significance level of a regression problem: regressions with losses lower than this threshold, with at least a certain probability, allow us to conclude that there is enough evidence of a linear relationship in the population. Under $H_0$ (correlation level equal to zero), the expected loss can be computed on a mesh-grid of uniformly distributed points ($\mathbf{x}$, $y$) within a standardized hypercube of $P+1$ dimensions centered at the origin. After some calculus with $P=1$, we find (see appendix) $\mathcal{R}=\frac{a^2}{6b}$ for $\mathcal{L}_1$ and $\mathcal{R}=\frac{b^2+a^2}{3}$ for $\mathcal{L}_2$, where $a$ and $b$ are the maximum values for $\hat{y}$ and $y$, respectively. These values can be easily approximated under $H_0$, as illustrated in figures \ref{fig:1c} and \ref{fig:1e}. It is worth mentioning that when using the $\mathcal{L}_1$ loss function, widely employed in predictive models for linear regression, such as K-fold and leave-one-out CV, the results are overly optimistic. This is because they provide convergent values lower than the theoretical expected loss under $H_0$.

\subsection{Classical tests of hypothesis in a linear model}
A learning algorithm fits a linear function $f(x) \equiv \hat{y}= \beta_1 x + \beta_0$ using a loss function that penalizes the difference between the observation and prediction, e.g. the OLS algorithm. A classical test for linearity on the OLS estimates conducts a test of hypothesis about the regression parameter $\beta_1$, where errors are assumed to be independent random quantities normally distributed with mean zero and a common variance. The sampling distributions of the OLS estimates $\hat{\beta}_i$ are indeed normal and a suitable test statistic for testing the null distribution on individual $\beta_i$'s $H_0: \beta_i=0$ against the alternative hypothesis $H_1: \upbeta \neq 0$, is the t-test: 
\begin{equation}
  t=\frac{\hat{\beta}_i}{std(\hat{\beta}_i)}  
\end{equation}

In addition to examining individual hypotheses by the t-test, the classical F-test is defined to test $P$ multiple hypothesis given $N$ observations. It is defined as the ratio 
\begin{equation}
    F = \frac{MSR}{MSE}
\end{equation}
with $P$ and $N-P-1$ degrees of freedom, where $MSE=\sum (y_i-\hat{y}_i)^2/(N-P-1)$ is the mean square error and $MSR=\sum (\hat{y}_i-\bar{y})^2/P$ is the mean square due to regression \cite{Chatterjee06}. 

After linearity is confirmed at a significance level $\alpha$ the quality of the fit can be measured by the magnitude of the t/F-test, the correlation coefficient $Cor(y,x)$, coefficient of determination $R^2= Cor(y,\hat{y})$, etc.; however they all require the aforementioned assumptions. In the following sections we show how the empirical risk, described as a functional area, can be used as a test for linearity and to assess the quality of the fitting, without the assumptions stated earlier. 

\section{A Test for Linearity Using Error Estimation in ML Algorithms}\label{sec:test}

In this section, we present a non-parametric method for testing the null hypothesis utilizing a common measure employed by ML researchers, namely the expected loss $\mathcal{R}$. Under $H_0$, we expect the loss value obtained with the linear regressor $f_N \in \mathcal{F}$ to be comparable to that of uncorrelated variables. Conversely, this value is expected to be lower when the regression coefficients are significant. The power of the test ($1-\beta$), control of false positives (FPs) and more statistical properties will be empirically assessed in the experimental section. 

\subsection{The SAR test}
The proposed non-parametric test, named the Statistical Agnostic Regression (SAR) test, evaluates the significance level $\alpha$ used in the regression analysis by comparing the actual risk with that obtained under the assumption of no correlation between predictors and observations. At a significance level $\alpha$, the non-parametric SAR test for linearity is formulated as follows:
\begin{equation}\label{eq:cinco}
\begin{array}{c}
H_0: \upbeta = 0 \\
H_1: \upbeta \neq 0  
\end{array}
\end{equation}
In this framework, the test statistic is $\mathcal{R}(f_N)$, and it is calculated with at least a probability of $1-\eta$ by considering the \emph{worst-case scenario} as:
\begin{equation}\label{eq
} \mathcal{R}(f_N) = \mathcal{R}_N(f_N) + \Delta(N, \mathcal{F}) \end{equation}
where we incorporate the empirical risk $\mathcal{R}_N(f_N)$ and an upper bound $\Delta(N, \mathcal{F})$. We reject $H_0$ if the $\mathcal{R}$ statistic, computed with at least a probability of $1-\eta$, is less than the critical value $\gamma$, or if its p-value, derived after randomization, is less than the level of significance $\alpha$. Otherwise, we fail to reject it.

Under $H_0$, the critical value $\gamma$ is equal to the expected loss when $\upbeta=0$; for example, $\gamma=\frac{1}{N}\mathbf{y}^T\mathbf{y}$ for $\mathcal{L}_2$, similar to the perfect orthogonality in OLS. For comparison purposes, we propose an extension of the permutation test, where the threshold $\gamma$ varies with the sampling process (bootstrapping). We determine how likely the evidence for linearity is by resampling the available data and using the empirical risk distribution, the observed statistic $\mathcal{R}$, compared to the critical value from $R$ realizations. The probability of the observed value of the expected loss under $H_0$ through permutation $\pi$ is:
\begin{equation}\label{ec:pvalue}
p_{value}=\frac{\#(\mathcal{R^\pi}\geq\gamma)}{R+1}
\end{equation}
where $\#(.)$ represents the number of times the risk in the permutation is greater than or equal to the error obtained under the null hypothesis $H_0$.  If this p-value is less than our level of significance, e.g. $\alpha=0.05$, then there is evidence to reject $H_0$.

The former test is similar to the aforementioned F-test (also known as analysis of variance). The difference lies in the model assumptions, e.g., in classical approaches we assume $E[MSE] = \sigma^2$, and in the methods used to estimate the expected losses. If the aforementioned assumptions are fulfilled, the $F^*$  ratio transforms into an $F$ distribution, and the regression problem becomes a test of hypothesis or the analysis of p-values. Even if the linear model assumptions are not fulfilled, we can test our ML measures in the same manner \cite{Reiss15} and compare them with the test provided in equation \ref{eq:cinco} (see the experimental section). Moreover, previous approaches to achieve statistical significance using ML and statistical agnostic theory in classification tasks \cite{Gorriz23} are naturally extended in this regression analysis.

\section{Materials and Methods}\label{sec:matmet}

\subsection{A Probably Approximately Correct Bayesian bound}

In this work, we leverage a significant advancement in the field rooted in Probably Approximately Correct (PAC)-Bayesian theory \cite{MacAllester2013}. Specifically, we assess a dropout bound inspired by the recent success of dropout training in deep neural networks. The bound, as represented in equation \ref{eq:dos}, is articulated with respect to the underlying distribution $Q$, which samples the function $f$ from the set of 'rules' denoted as $\mathcal{F}$.

For any constant $\lambda>1/2$ and a class of linear regressors, $\upbeta\in \mathbb{R}^{P+1}$, selected according to the distribution $Q$, we establish that with a probability of at least $1-\eta$ over the sample draw, the following CI are valid for all distributions $Q$ with dropout rate $\delta$:
\begin{equation}
\begin{array}{l}
\mathcal{R}(\upbeta)\leq \mathcal{R}_N(\upbeta) + \\
 \min_{1\leq i\leq k}\frac{1}{2\lambda_i-1}\left(\mathcal{R}_N(\upbeta)+
\frac{2\lambda_i^2\mathcal{L}_{max}}{N}\left(\mathcal{D}(Q,Q_u))+\ln\frac{k}{\eta}\right)\right)
\end{array}
\end{equation}
Here, $\mathcal{D}(Q,Q_u)=\frac{1-\delta}{2}||\boldsymbol{\upbeta}||^2$ represents the Kullback-Leibler divergence from $Q$ to the uniform distribution $Q_u$ that is formalised as an isotropic unit-variance prior $\mathcal{N}(0,1)^{P+1}$, $\lambda\in (1/2,10)$ can assume $k$ different values and $\mathcal{L}_{max}$ is an outlier threshold  \cite{MacAllester2013}. Other approaches to formulate upper bounds can be considered based on more general assumptions \cite{Vapnik82,Gorriz19}.

\subsection{ANOVA and the analysis of residuals in ML}
\label{sec:Anova}

In classical regression \cite{Chatterjee06} results are tested for significance by assessing the analysis of variance of the residuals $r=y-\hat{y}$.  As aforementioned, the $F^*$-statistic is compared to an F-distribution with $P=1$ numerator and $N-2$ denominator degrees of freedom and the p-value, or the probability that we get this statistic as large as we did under the null-hypothesis, is determined. This is also known as the formal test for the slope parameter $\upbeta_1$ and, if all the model assumptions are fulfilled, can be extended to evaluate the residuals obtained by ML methods.

Another possibility for drawing conclusions about the population (and not only the particular observed sample) is by the use of confidence intervals. Confidence intervals and hypothesis tests are two different ways of learning about the values of population parameters ($\upbeta_0$ and $\upbeta_1$). Both approximations can be conducted on ML residuals as well to demonstrate the reliability of the results, however this is rarely found in the ML literature and there is a trend to exclusively show different performance scores based on empirical measures; e.g. accuracy in learning curves. Similar to the permutation analysis proposed for group comparisons \cite{Bullmore99}, we can test the power $(1-\beta)$ of the proposed SAR test by comparing a set of Monte Carlo simulations with the expected loss under $H_0$, following the ideas presented in previous sections. We estimate the power $(1-\beta)$ by counting the number of times the corrected empirical loss was less than the averaged loss when $\upbeta=0$.

\subsection{Heteroscedascity and the Breusch-Pagan test}
\label{sec:BPtest}

Beyond linearity, we explore the concept of heteroscedascity \cite{Breusch79} using the ML approaches vs. OLS. When the residuals are heterogeneously distributed along the explanatory variable (predictor), we encounter the issue of heteroscedasticity in the data. The Breusch-Pagan (BP) test checks the null hypothesis (homoscedasticity) by re-fitting the explanatory (independent) variable to the squared residuals as observation variables. A simple way to assess this condition is by using the test statistic $T=N\cdot R^2$, where $R$ is the coefficient of determination of the auxiliary fitting, and follows a $\chi^2$ distribution with $P$ degrees of freedom under $H_0$ \cite{Koenker81}. If the p-value associated with it is less than the level of significance, we reject $H_0$. The homoscedasticity condition is assumed in the linear regression model, and when it fails, further analyses based on the F-test are not only imprecise (biased) but also invalidated from a statistical point of view. ML residuals could be tested in the BP test as well, to assess the ability of data-driven models to detect heteroscedasticity in regression problems with limited sample sizes.

\subsection{Synthetic, realistic  \&  real datasets}
\label{sec:datasets}

\begin{figure}
\centering
\includegraphics[width=\textwidth]{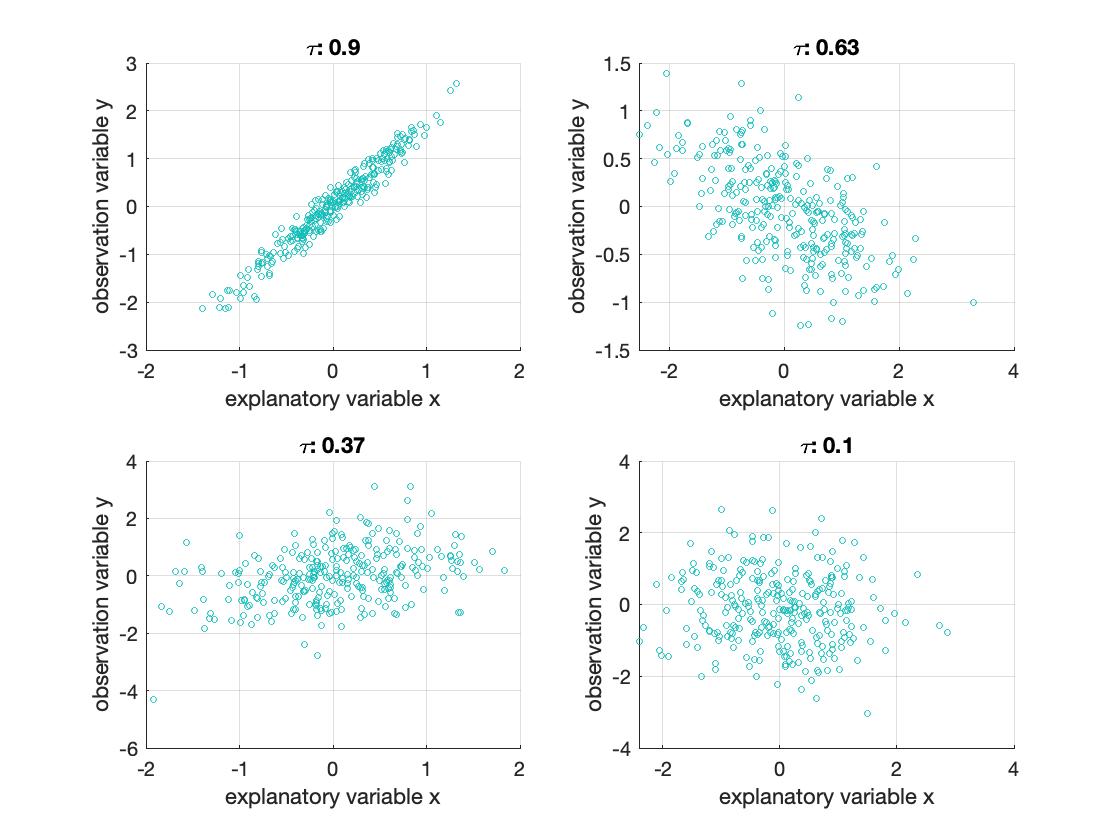}
\caption{Data transformed by rotation and scaling with a non-diagonal covariance matrix, assuming a Gaussian distribution.} 
\label{fig:dataG}%
\end{figure}

We evaluated linear SVR models using the validation techniques: K-fold, LOO, Resubstitution, and SAR, along with the OLS method, on various synthetic and realistic datasets. Note that no hyperparameter optimization was applied to the linear ML models; thus, overfitting is not expected, and nested cross-validation is not necessary in this experimental setup. First, we assessed the 2D problem with Gaussian-distributed variables ranging from perfectly uncorrelated to strongly correlated signals. To control the degree of correlation, we drew samples $\mathbf{Z}$ from a 2D normal distribution, i.e., $\mathcal{N}(0,1)$. Subsequently, we computed a random linear transformation matrix $\mathbf{T}$ and modified its singular value decomposition (SVD) diagonal matrix using a scaling transform similar to the earlier simulated dataset (\ref{fig:example1}). Finally, we applied this transformation matrix to the original data, obtaining $\hat{\mathbf{Z}}=\mathbf{Z}\mathbf{T}$ that generated observations and predictors (see figure \ref{fig:dataG}). We fitted our models on this data with two losses $\mathcal{L}_1$ and $\mathcal{L}_2$,  several samples sizes $N=10,20,\ldots,300$ and tested for significance using the F-test and the SAR test. We drew up to $R=100$ realizations (sampling) from the ideal distribution and then averaged the regression results to study the effect of sample size \cite{Friston13}. Whenever CV methods split the dataset into folds, we analyzed the variability of the performance measures accordingly. This represents a realistic situation with a fixed realization or dataset instead of having an ensemble of $R$ realizations. 

\begin{figure}
\centering
\includegraphics[width=\textwidth]{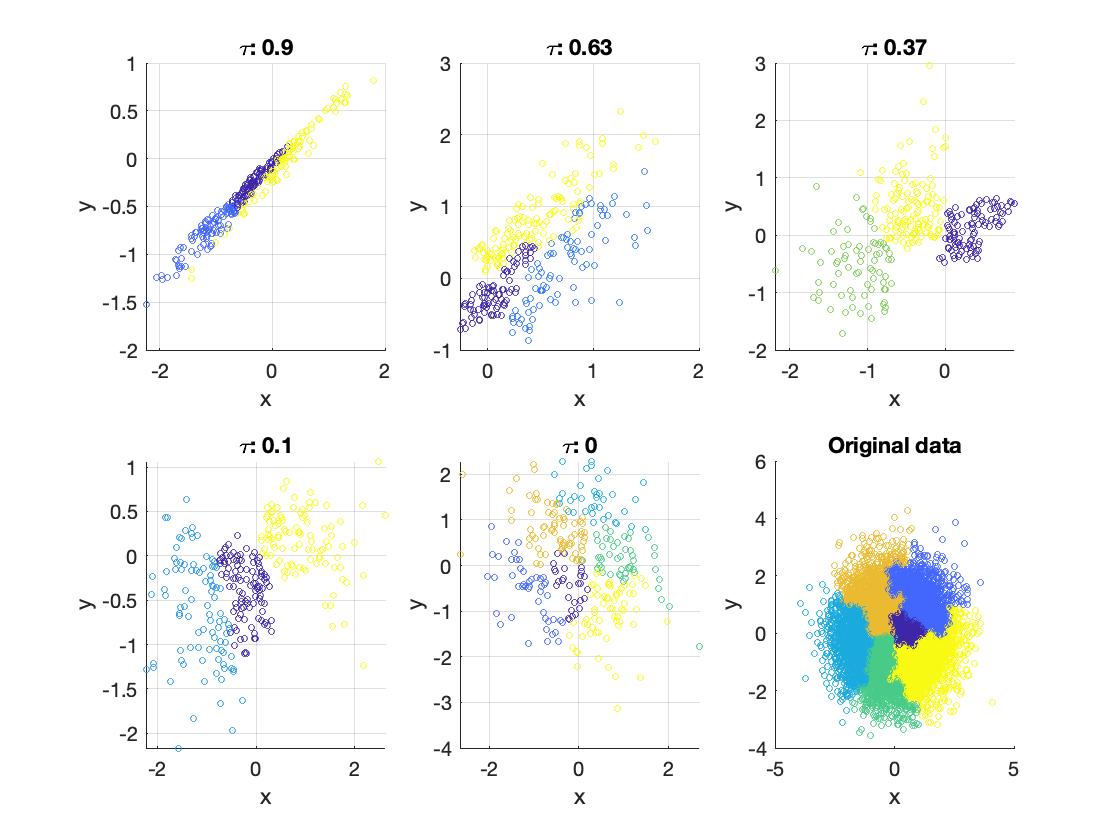}
\caption{Gaussian data transformed by rotation and scaling, along with cluster pruning. Colors simply indicate the applied transformations to the data and identify the clusters that were removed. } 
\label{fig:datanG}%
\end{figure}

To enhance the realism of our data, we employed the procedure used in \cite{Gorriz19} to segment the data into $N_c$ clusters. By selectively removing some of these clusters, we introduced non-Gaussian characteristics. Under this experimental setting, the F-test and, in particular, the OLS method are no longer optimal. This allows us to truly test the robustness of the SAR test developed in Section \ref{sec:background}. We conducted tests for up to four experimental settings by randomly removing a set of $N_c - N_g$ clusters, with $N_g = 3$. In figure \ref{fig:datanG} we plotted the remaining clusters with different colors to illustrate the complete data generation.

\begin{figure*}
\centering
\begin{subfigure}{0.49\textwidth}
    \includegraphics[width=\textwidth]{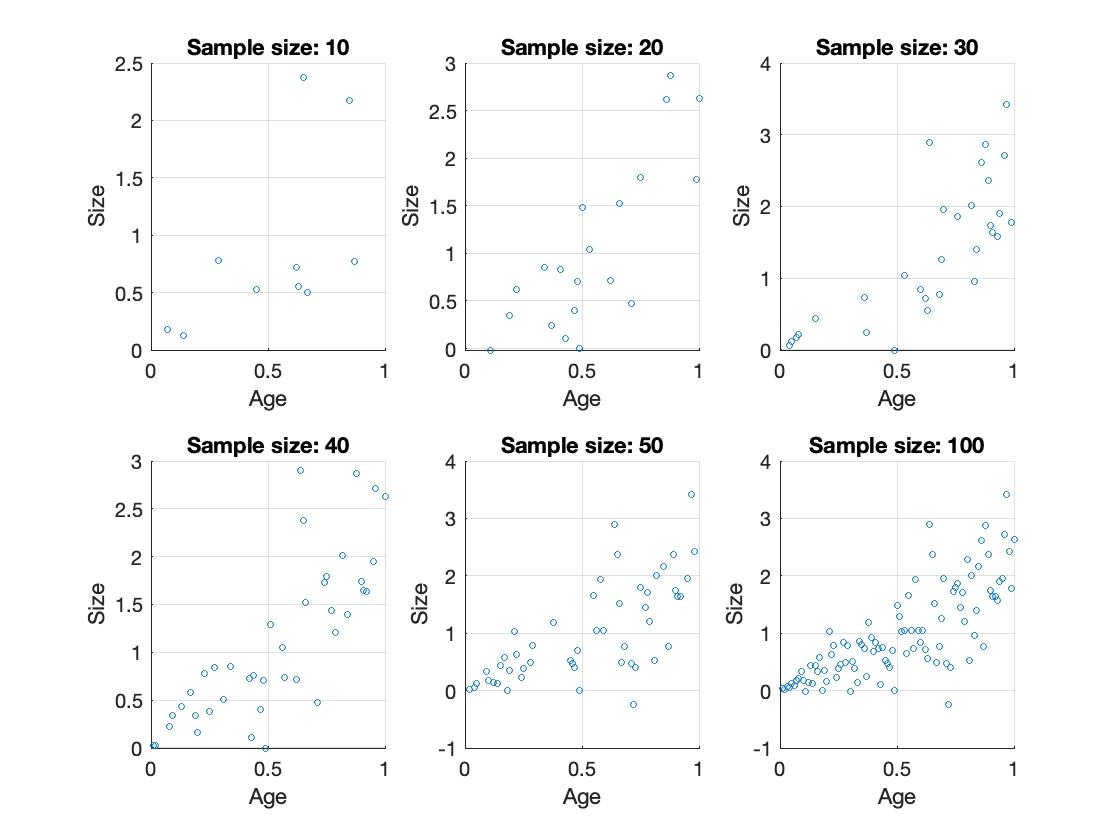}
    \caption{Datasets.}
    \label{fig:5a}
\end{subfigure}
\hfill
\begin{subfigure}{0.49\textwidth}
    \includegraphics[width=\textwidth]{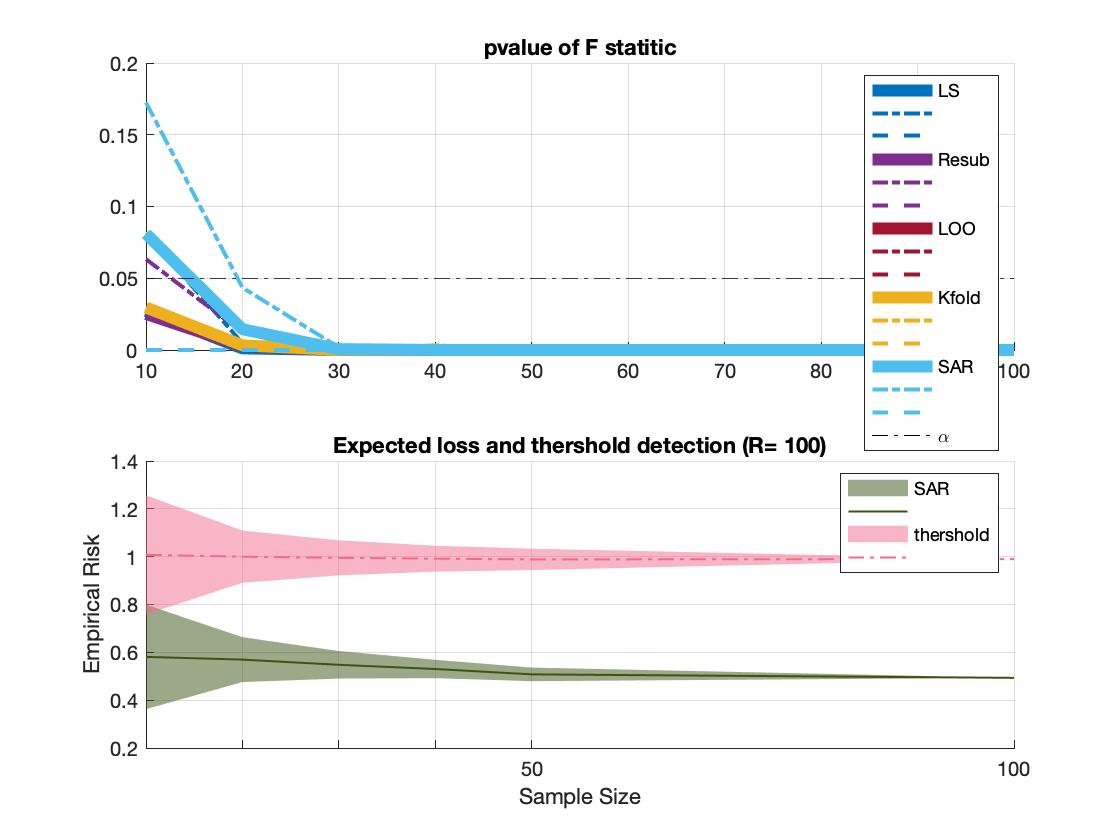}
    \caption{Tests for linearity.}
    \label{fig:5b}
\end{subfigure}
\hfill
\begin{subfigure}{0.49\textwidth}
    \includegraphics[width=\textwidth]{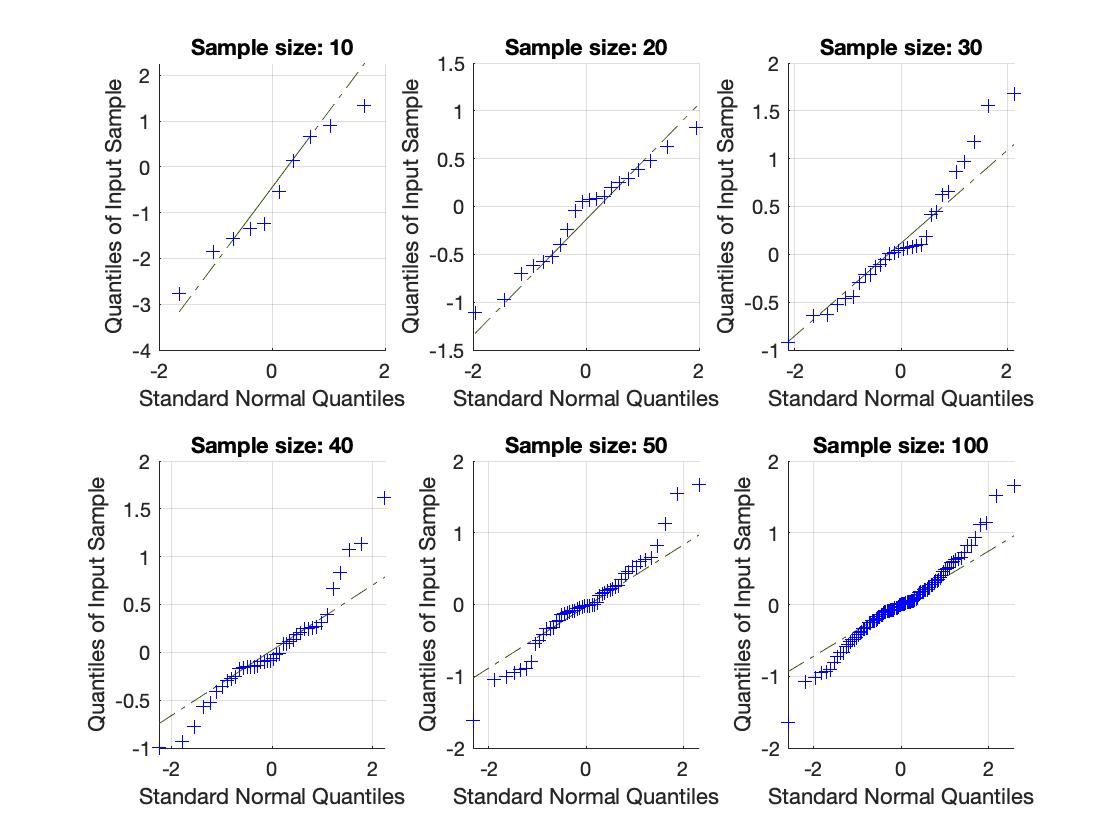}
    \caption{Q-Q plot revealing non-Gaussianity.}
    \label{fig:5c}
\end{subfigure}
\hfill
\begin{subfigure}{0.49\textwidth}
    \includegraphics[width=\textwidth]{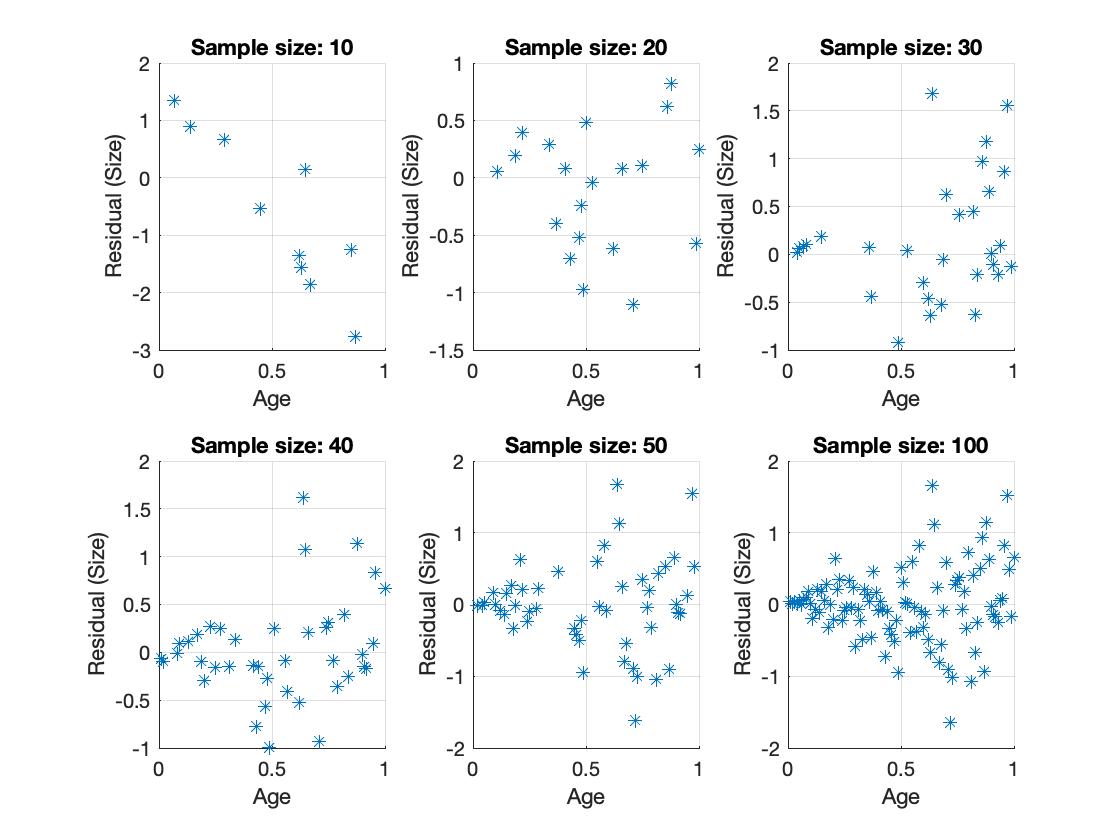}
    \caption{Explanatory variable vs. residuals.}
    \label{fig:5d}
\end{subfigure}
        
\caption{Dataset with heteroscedascity and increasing sample size in figure \ref{fig:5a} and SAR and F tests on linearity in figure \ref{fig:5b}.  Assumptions needed to perform the F-test are not fulfilled as shown in the Q-Q plots in figure \ref{fig:5c} and residuals vs. explanatory variable plot in figure \ref{fig:5c}.}
\label{fig:BP1}
\end{figure*}

Finally, we assessed how the BP test worked on the residuals obtained from the set of methods previously described. Specifically, we evaluated the ability of the resulting models to detect heteroscedasticity versus sample size. For this purpose, we employed the dummy dataset provided in the XLSTAT software \cite{Soft19} designed to compare a homoscedastic model to another with strong heteroscedasticity. In particular, we generated a dependent variable (``Size'') based on an independent variable (``Age''), where the residuals are defined as the product of the independent variable by the random normal error (figure \ref{fig:5a}). In this case, the residuals were obviously correlated with age, and the problem (figure \ref{fig:BP1}) was apparently linear, but heteroscedastic.

\subsubsection{A neuroimaging dataset to study Alzheimer Disease}

Data used in preparation of this paper were obtained from the Alzheimer's Disease Neuroimaging Initiative (ADNI) database (adni.loni.usc.edu). The ADNI database contains 1.5 T and 3.0 T t1w MRI scans for AD, Mild Cognitive Impairment (MCI), and cognitively NC which are acquired at multiple time points. Here we only included 1.5T sMRI corresponding to the three different groups of subjects. The original database contained more than 1000 T1-weighted MRI images, comprising $229$ NC, $401$ MCI (252 stable MCI and 149 progressive MCI) and $188$ AD, although for the proposed study, only the first medical examination of each subject is considered, resulting in $N=818$ segmented GM images after standard preprocessing. Demographic data of subjects in the database is summarized in Table \ref{tab:demog}. 

\begin{figure*}
\centering
\begin{subfigure}{\textwidth}
    \includegraphics[width=\textwidth]{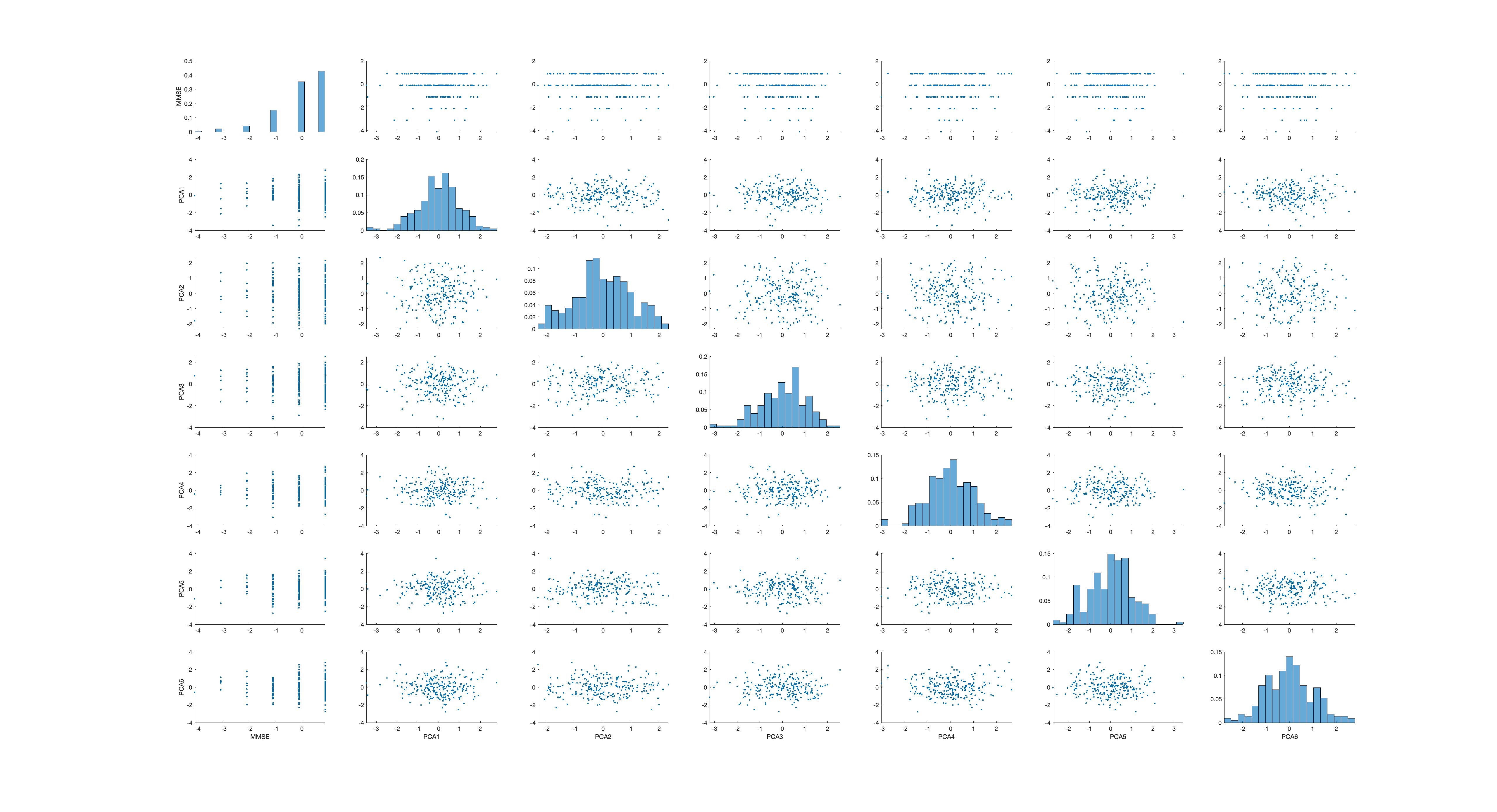}
    \caption{Scatter plots and histograms.}
    \label{fig:ADNIa}
\end{subfigure}
\hfill
\begin{subfigure}{0.49\textwidth}
    \includegraphics[width=\textwidth]{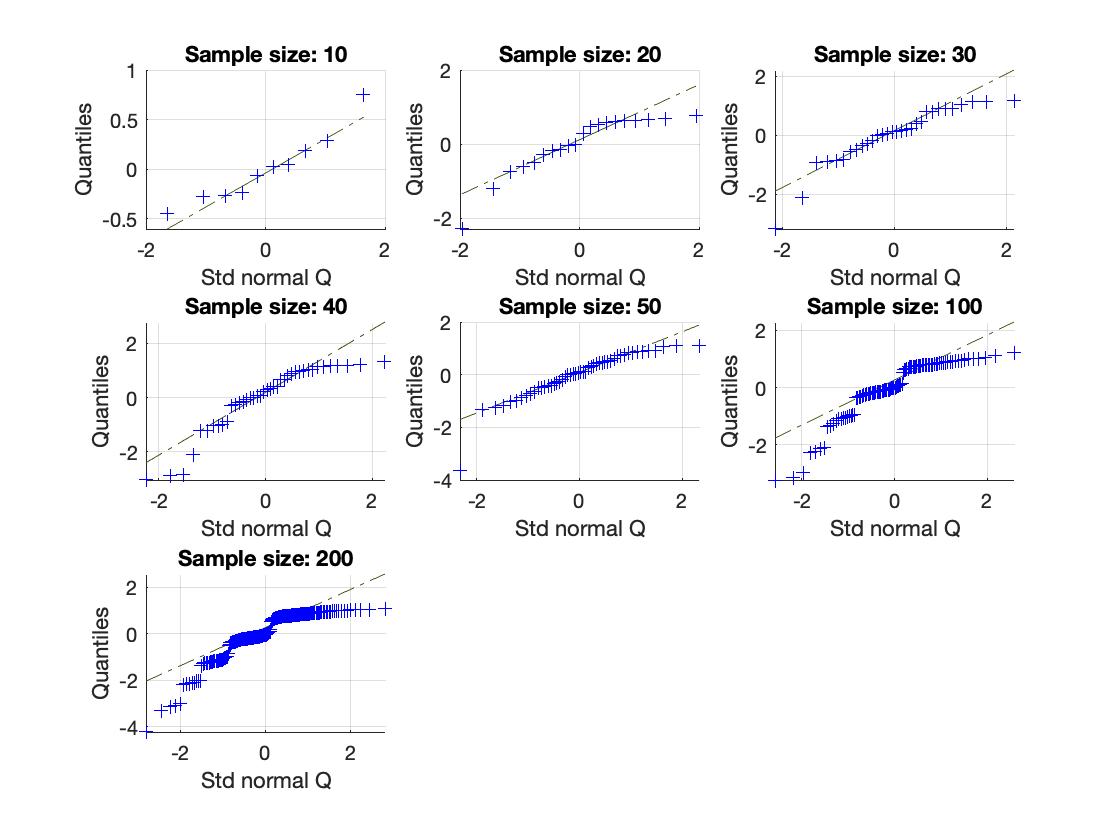}
    \caption{Q-Q plot revealing non-Gaussianity in the NOR class.}
    \label{fig:ADNIb}
\end{subfigure}
\hfill    
\begin{subfigure}{0.49\textwidth}
    \includegraphics[width=\textwidth]{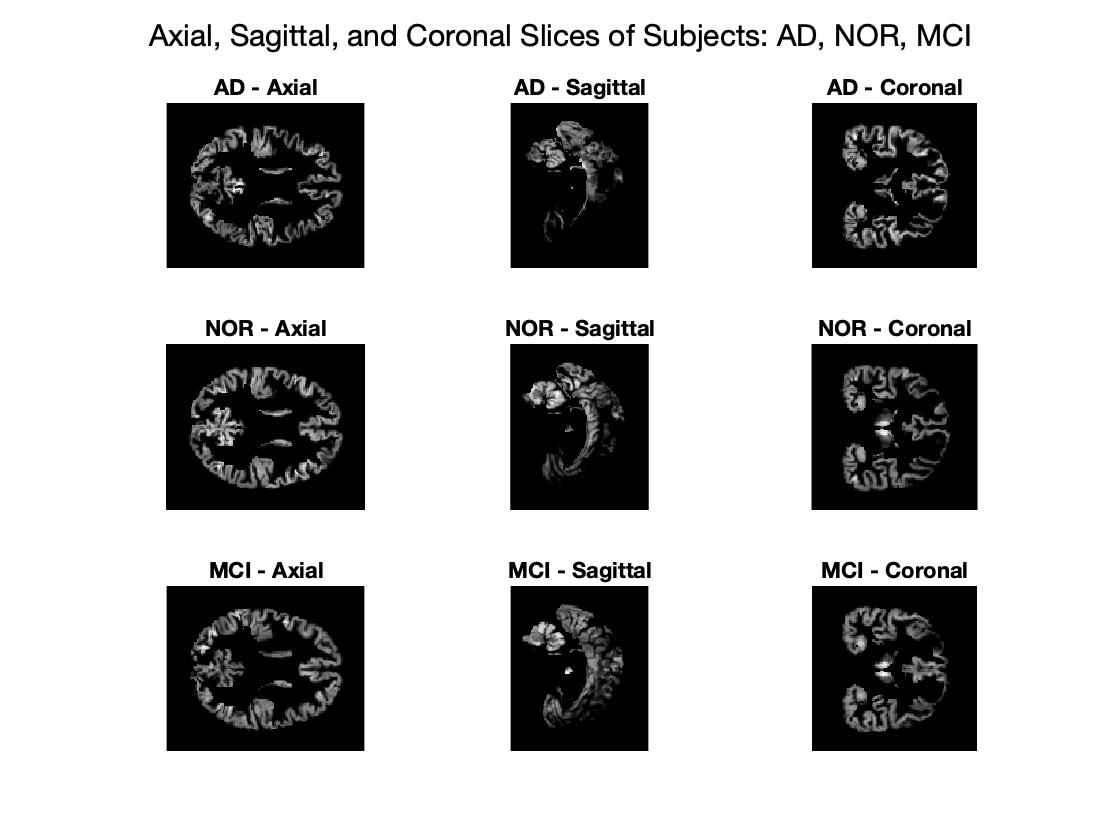}
    \caption{Three samples of the segmented GM dataset}
    \label{fig:ADNIc}
\end{subfigure}
\hfill    
\caption{ADNI Dataset with multiple predictors (6) in figure \ref{fig:ADNIa}. The first column represents the MMSE as the observable variable. In figure \ref{fig:ADNIb} we represent the Q-Q plot following the analysis given in the reference provided in the text. Assumptions needed to perform the F-test are not fulfilled as shown in the Q-Q plots in figure \ref{fig:6b}.}
\label{fig:ADNI}
\end{figure*}

Feature extraction based on principal component analysis (PCA) was performed on the set of segmented GM images to reduce high-dimensional data while retaining key variance and information. A simple pairwise scatter plot and histogram of these novel features for the NOR class with respect to demographic factors, such as the Mini Mental State Exam (MMSE), along with Q-Q plots of the corresponding OLS residuals, revealed non-Gaussianity and non-standard data distributions in the control class (figure \ref{fig:ADNI}).

\begin{table}[htbp]
   \centering
   \caption{Demographics details of the ADNI dataset with group means with their standard deviation}
   \label{tab:demog}
   \begin{tabular}{ccccc}
     \hline
     Status & Number &	Age	& Gender (M/F) &	MMSE\\ \midrule
 \\
NC	        &   229    &	  75.97$\pm$5.0	    &   119/110	  & 29.00$\pm$1.0 \\
MCI          &	  401   &	  74.85$\pm$7.4	    &   258/143	  & 27.01$\pm$1.8  \\
AD	        &  188	   &     75.36$\pm$7.5 	&     99/89	  & 23.28$\pm$2.0  \\
\bottomrule
\end{tabular}
\end{table}

\subsubsection{Cancer dataset in multiple dimensions}\label{sec:cancer}

\begin{figure*}
\centering
\begin{subfigure}{\textwidth}
    \includegraphics[width=\textwidth]{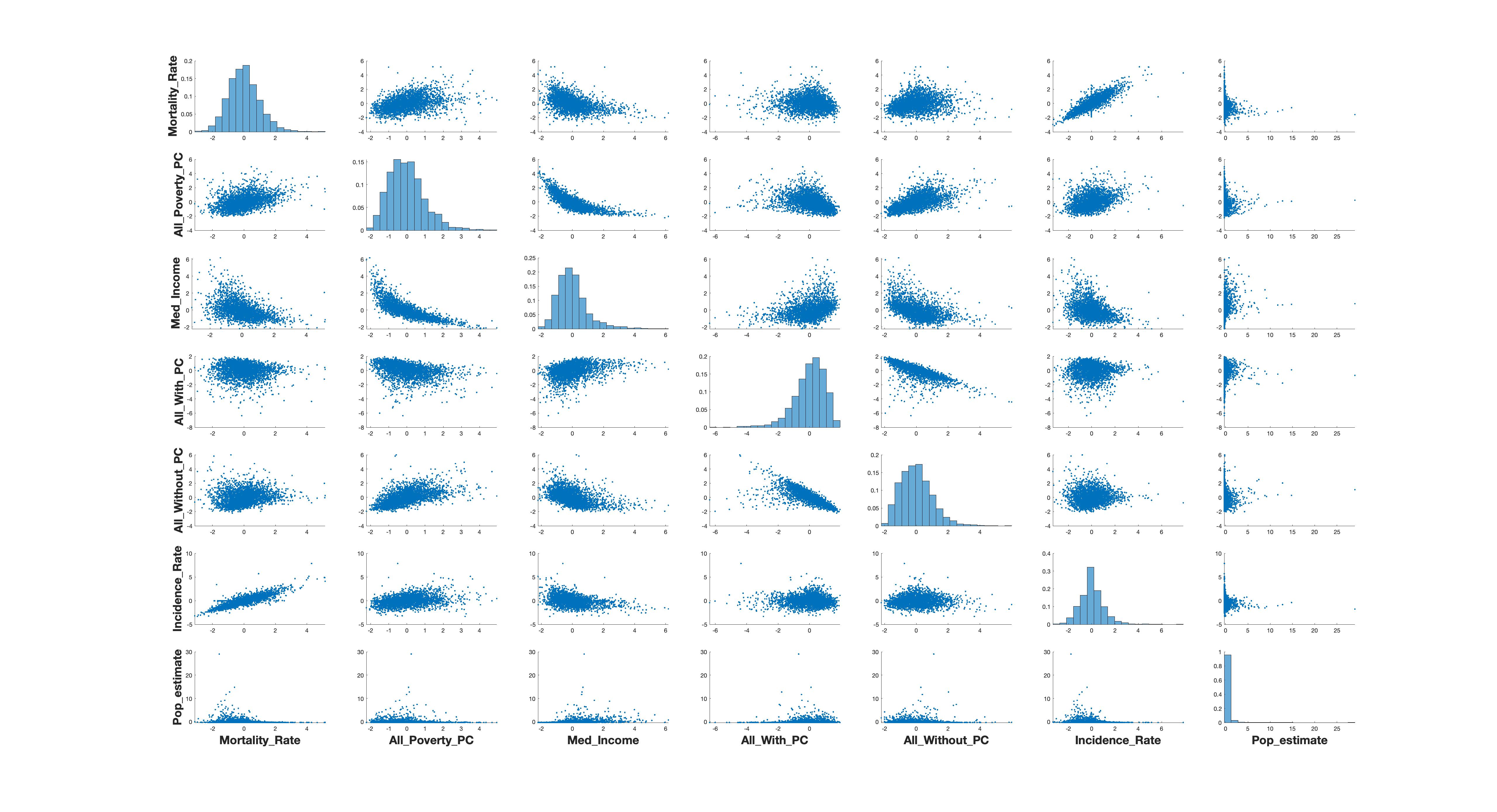}
    \caption{Scatter plots and histograms.}
    \label{fig:6a}
\end{subfigure}
\hfill
\begin{subfigure}{0.49\textwidth}
    \includegraphics[width=\textwidth]{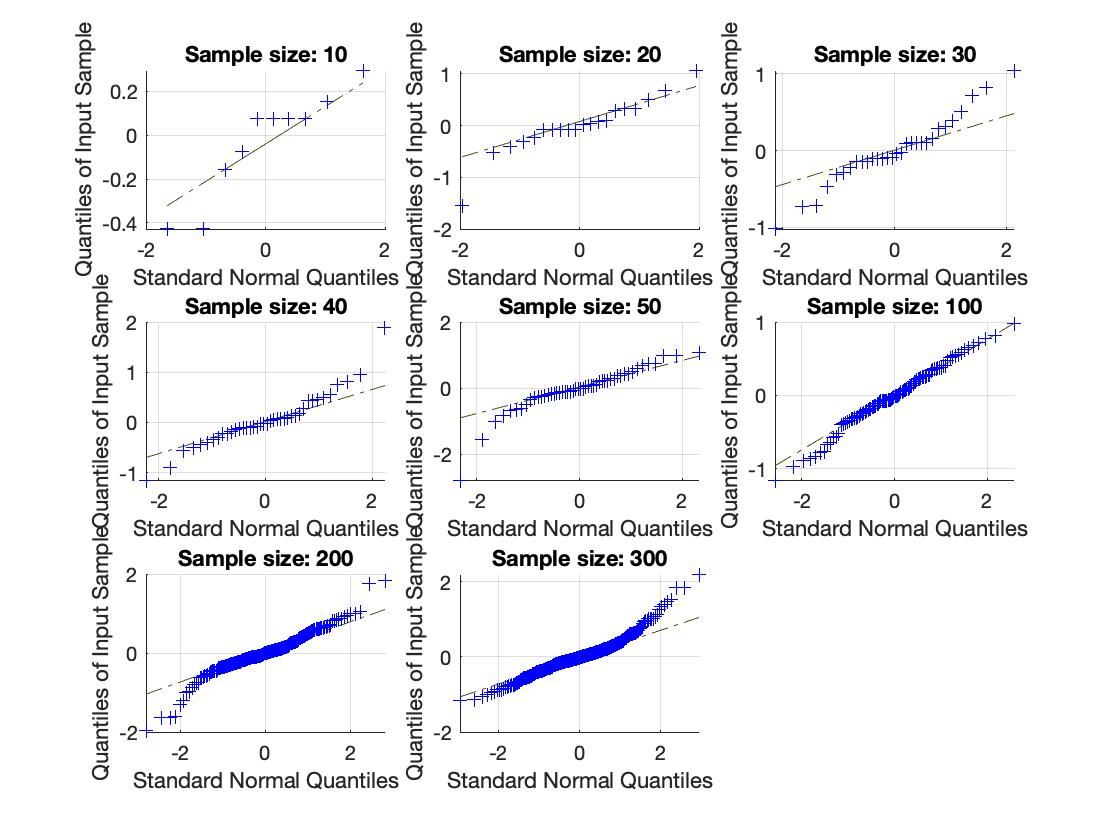}
    \caption{Q-Q plot revealing non-Gaussianity.}
    \label{fig:6b}
\end{subfigure}
\hfill
\begin{subfigure}{0.49\textwidth}
    \includegraphics[width=\textwidth]{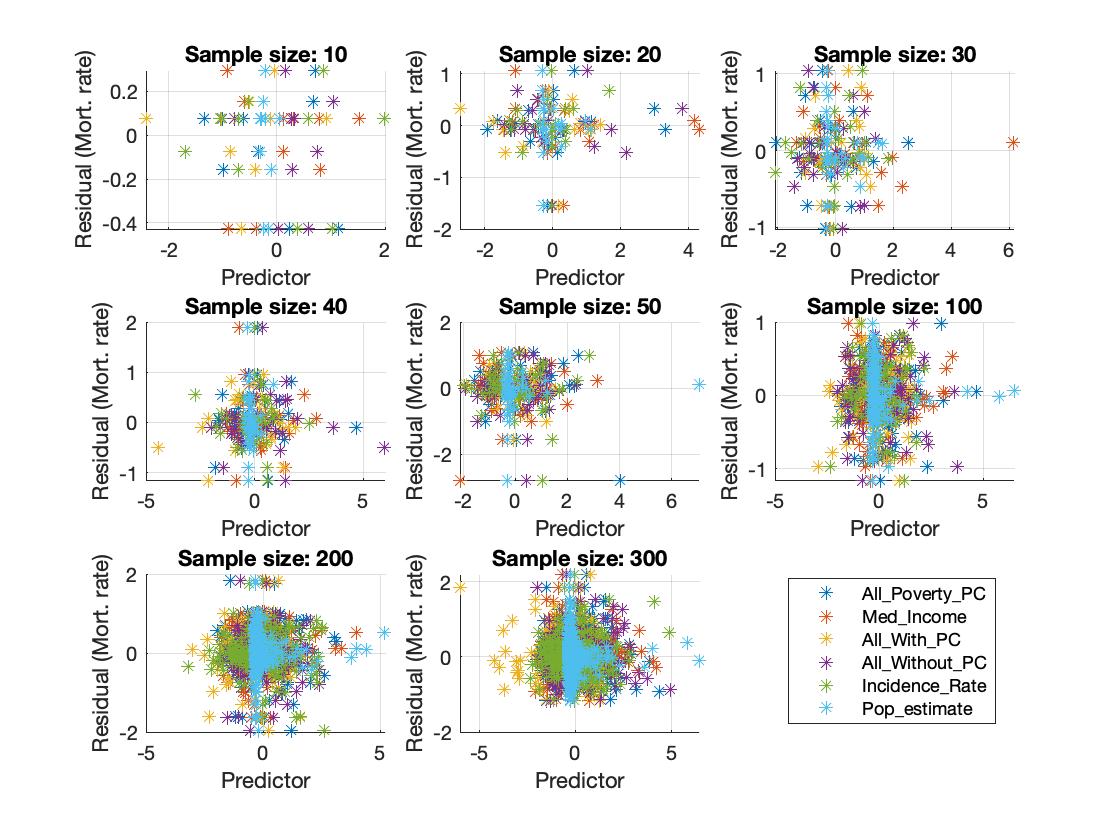}
    \caption{Explanatory variable vs. residuals.}
    \label{fig:6c}
\end{subfigure}
\caption{Cancer Dataset with multiple predictors (6) in figure \ref{fig:6a}. The first column represents the mortality rate as the observable variable. In figure \ref{fig:6b} we represent the Q-Q plot following the analysis explained in section \ref{sec:cancer}. Assumptions needed to perform the F-test are not fulfilled as shown figures \ref{fig:6b} and \ref{fig:6c}.}
\label{fig:cancer}
\end{figure*}

We tested our multivariate methods on real data set with data downloaded from US National Cancer Institute and the US Census American Community Survey to explore the linear relationship between between cancer mortality rate and several predictors (socio-economic status) in US counties. See \url{https://data.world/nrippner/cancer-linear-regression-model-tutorial/} for a full description of the data (see figure \ref{fig:6a}). 

Following the OLS analysis (figure \ref{fig:cancer}) of this dataset and eliminating multicollinearity by applying variance inflation factors, we employed up to six independent predictors including: Both male and female reported below poverty line per capita (All\_Poverty\_PC),  median income of all ethnicities (Med\_Income),  males and females with and without health insurance per capita (All\_With$/$out\_PC), lung cancer incidence rate (Incidence\_Rate), and the population estimate in 2015 (Pop\_estimate). A total of $N=2809$ sample size dataset was created and randomly down sampled for randomization analysis.

\section{Results}\label{sec:experiments}

\subsection{Gaussian data: OLS is the gold standard}

\begin{figure*}
\centering
\includegraphics[width=0.49\textwidth]{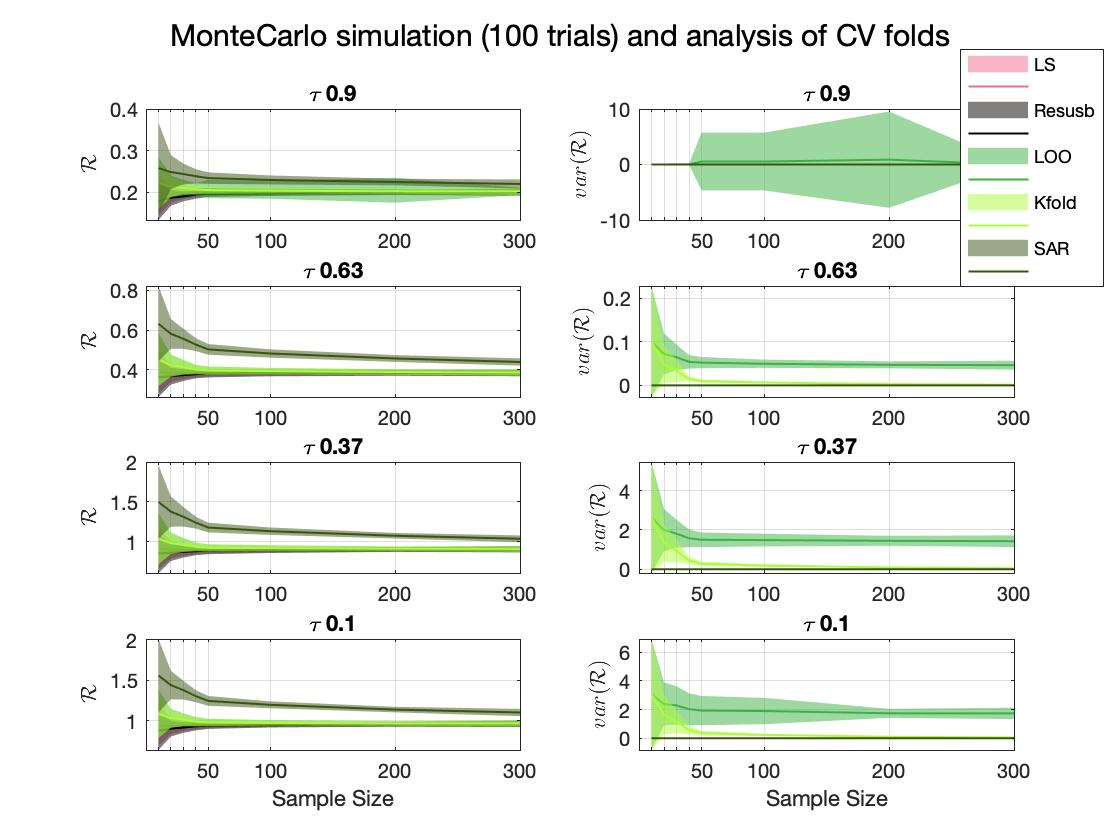}
\includegraphics[width=0.49\textwidth]{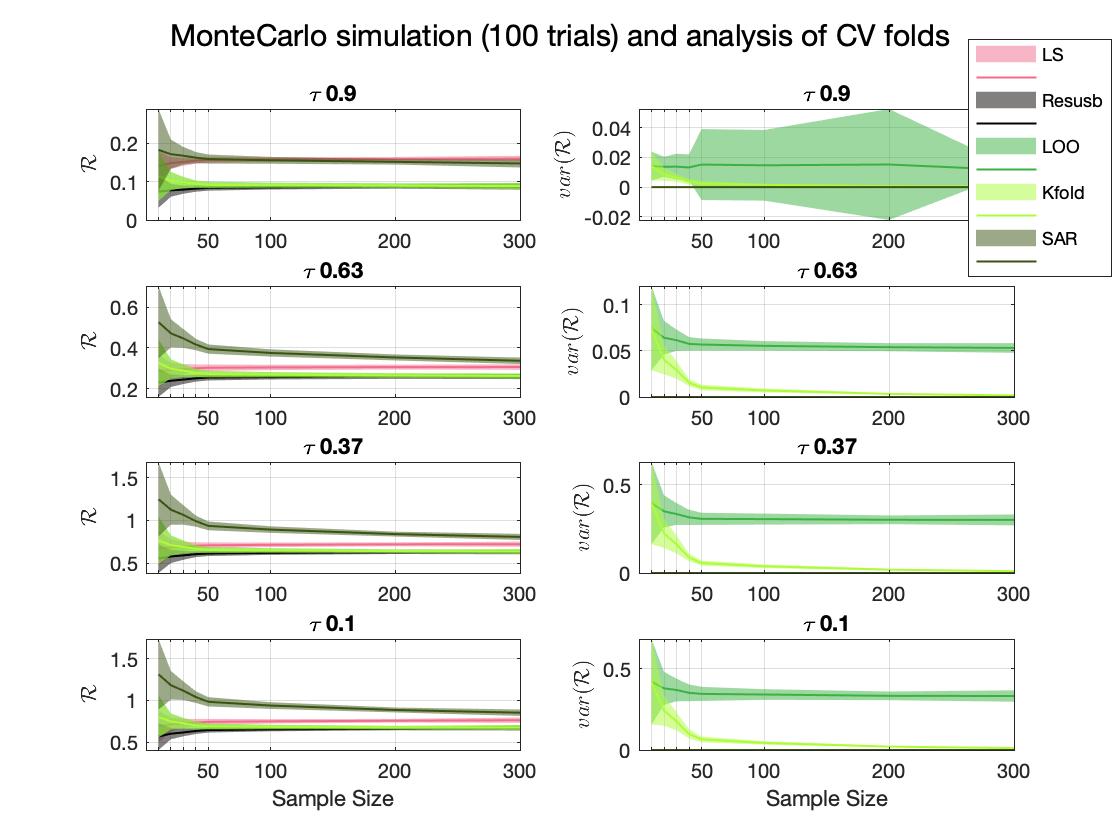}
\caption{Expected losses ($\mathcal{L}_2$ on the left and $\mathcal{L}_1$ on the right) and analysis of variance from folds. Note that the variance from folds is that experienced in real scenarios with a single sample realization.} 
\label{fig:dataGloss}%
\end{figure*}

\begin{figure}
\centering
\includegraphics[width=0.49\textwidth]{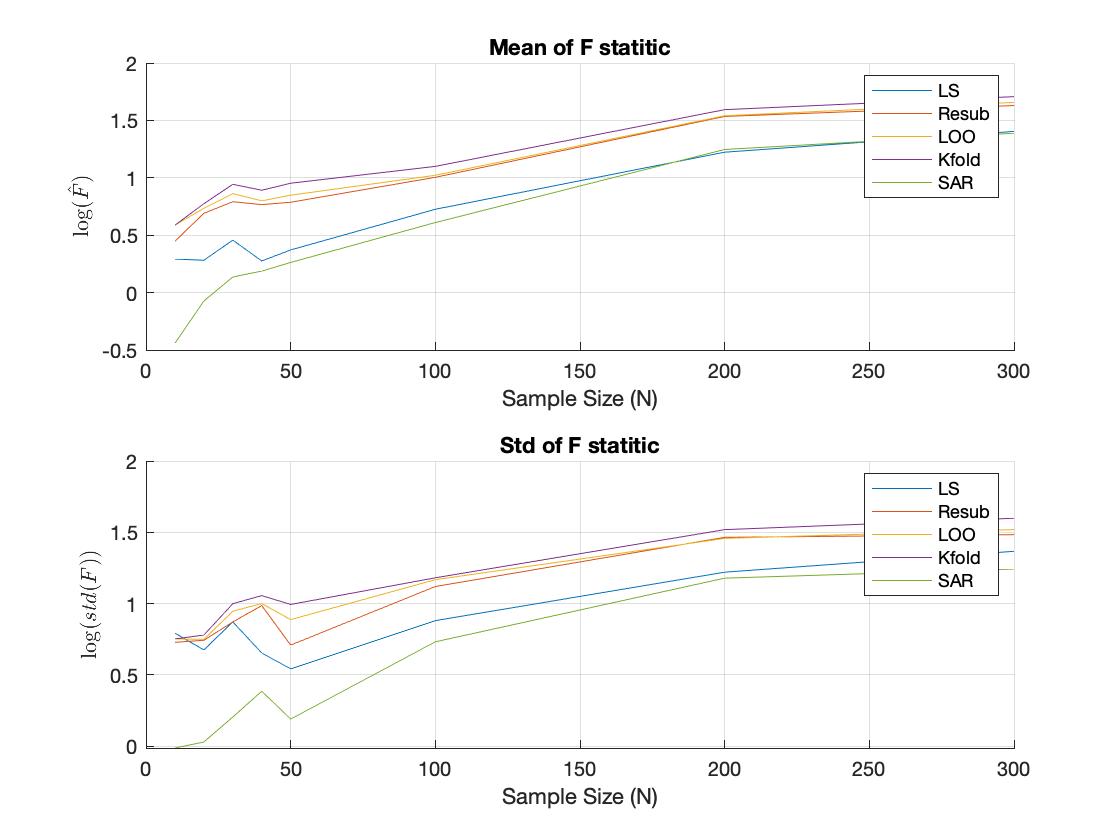}
\includegraphics[width=0.49\textwidth]{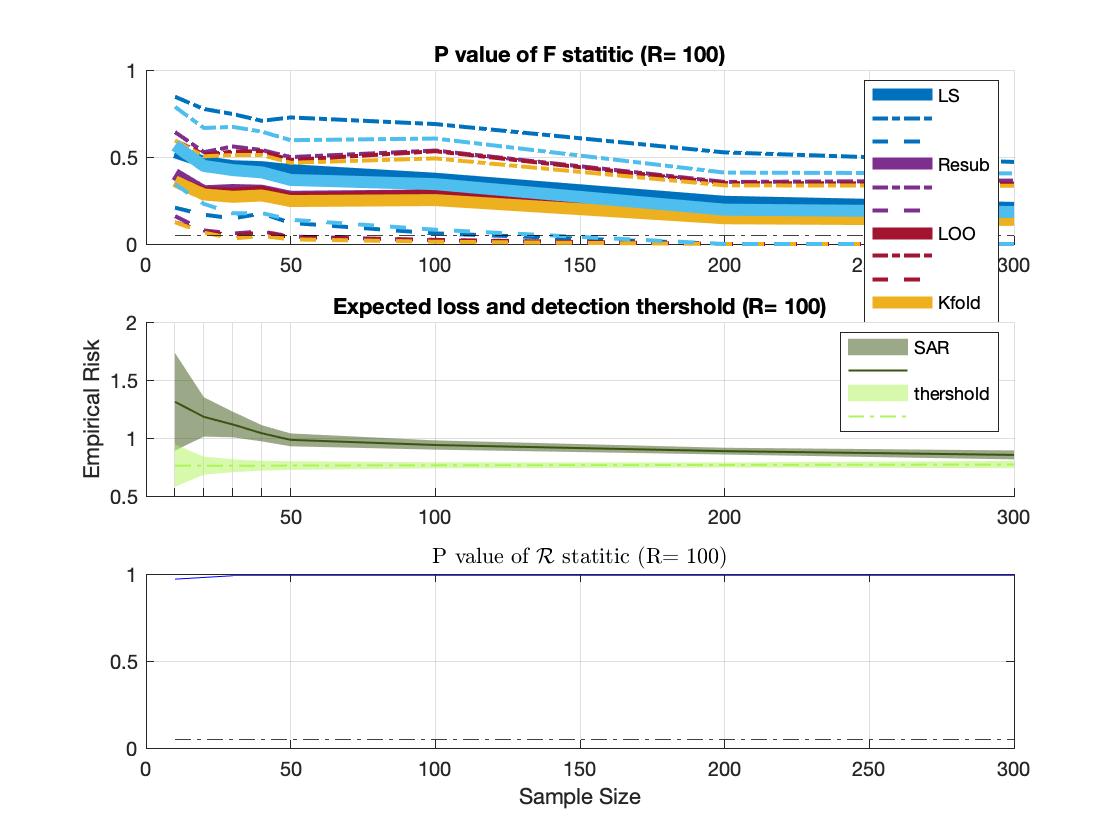}
\caption{The F-test for the slope $\upbeta_1$ in the case of a correlation level equal to 0.1. We utilized the residuals derived from each method and averaged the F-statistic and its variability over $R=100$ repetitions. It is noteworthy that SAR provides a trade-off between ML methods and OLS, exhibiting less variability in the repetitions. The SAR test is obtained with a probability of at least $\eta=0.5$.} 
\label{fig:Ftest}%
\end{figure}

\begin{figure}
\centering
\includegraphics[width=0.49\textwidth]{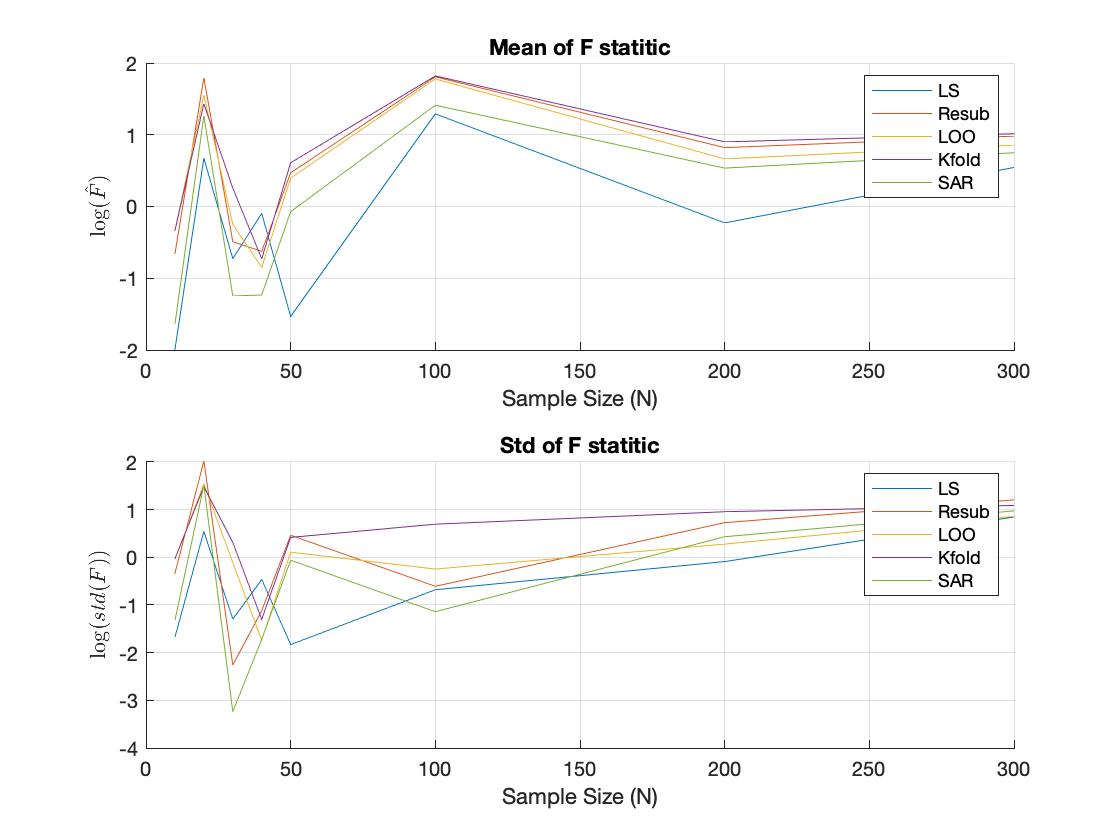}
\includegraphics[width=0.49\textwidth]{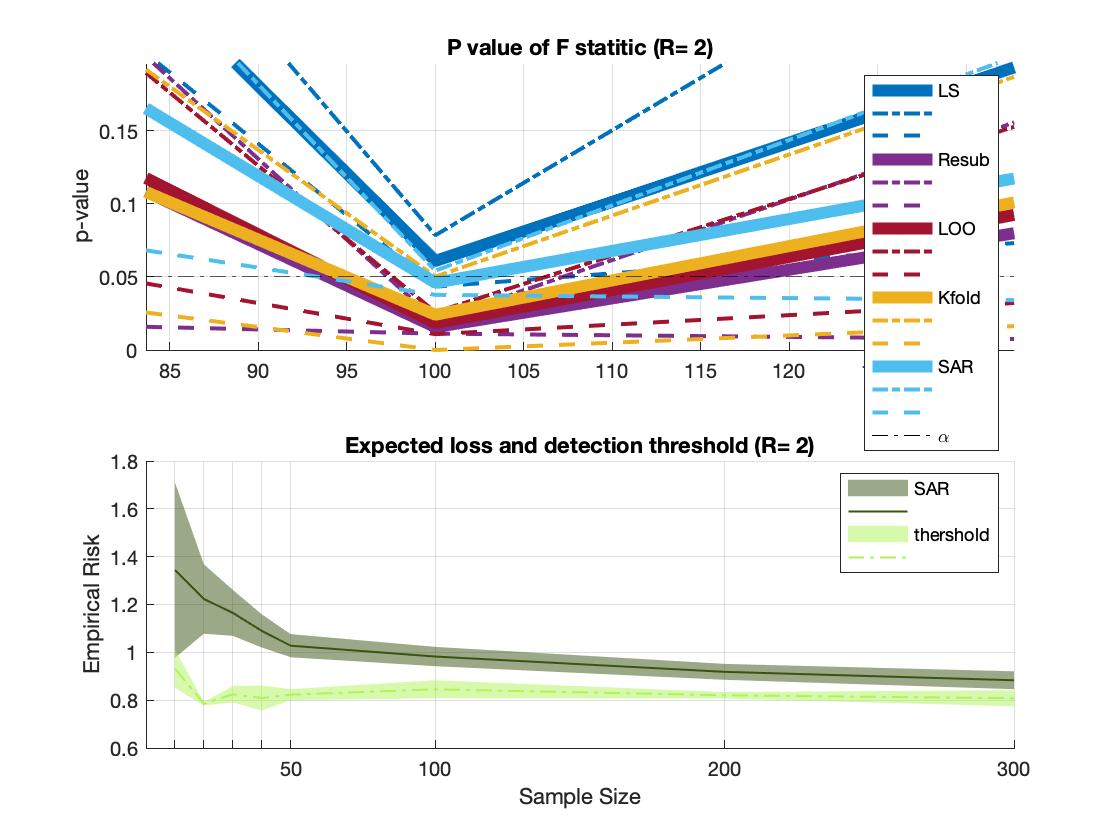}
\caption{The F-test for the slope $\upbeta_1$ in the case of a correlation level equal to $0.1$. We utilized the residuals derived from each method and averaged the F-statistic and its variability over $R=2$ repetitions.} 
\label{fig:Ftest2}%
\end{figure}

\begin{figure}
\centering
\includegraphics[width=\textwidth]{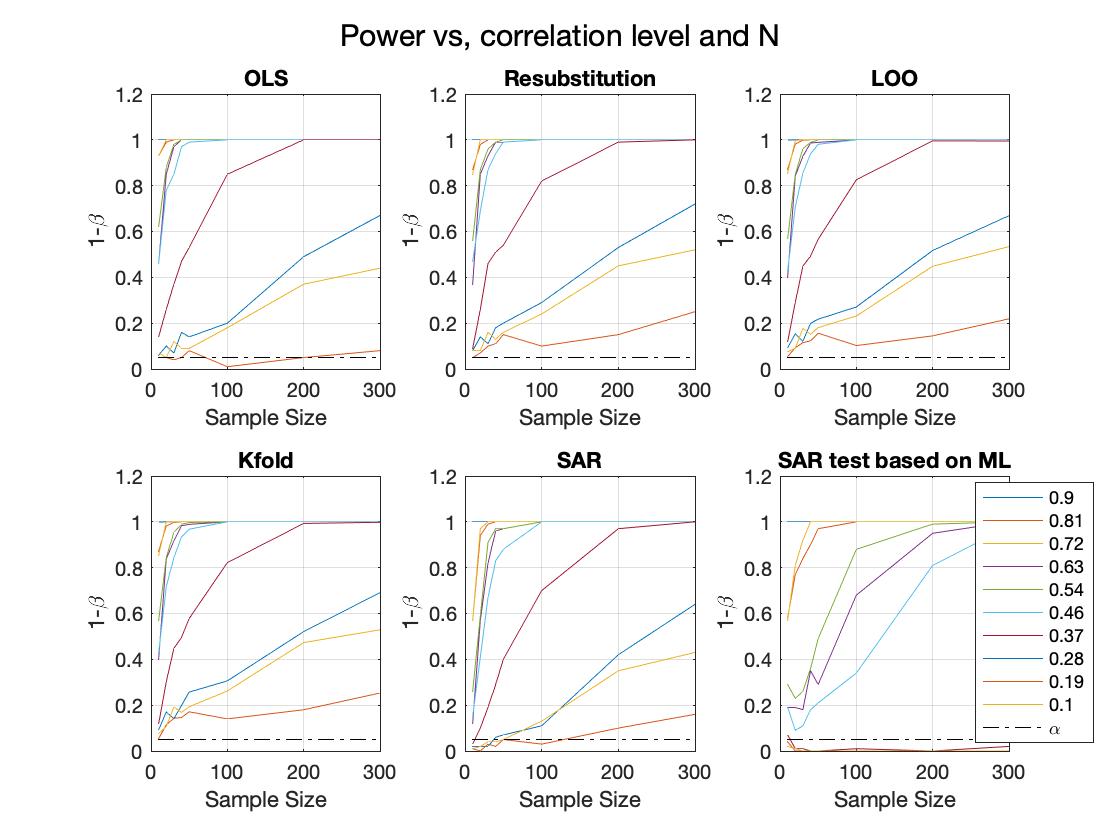}
\caption{Power analysis derived from the F-test in Gaussian data. The selected loss for ML methods is $\mathcal{L}_1$. Observe how the results obtained by the SAR features are by far the most similar to the ones obtained by the OLS. The SAR test based on expected loss is more conservative than the other methods (correlation levels below $0.37$ are not considered significant based on the worst-case analysis with a probability of at least $\eta=0.5$).} 
\label{fig:power1}%
\end{figure}

In this section, linearly transformed Gaussian data is considered, where the OLS method is the gold standard. We averaged $R=100$ realizations and used the same loss ($\mathcal{L}_2$) with standard ML algorithms, which performed quite well (SAR was conservative but converged to OLS as the sample size increased). However, there was high variability in the expected loss from folds when the number of samples was limited ($N<50$). It is worth noting that when using the $\mathcal{L}_1$ loss, the typical loss in ML applications, the correction applied to the resubstitution error made SAR converge to OLS, unlike the other ML validation methods (see figure \ref{fig:dataGloss}).
The methods are now tested with the formal F-test for the slope to assess their ability to detect a slight linear relationship with a correlation level equal to 0.1. The results reveal optimistic behavior of the ML methods, except for the SAR method which struck a trade-off between OLS and ML methods. It is essential to recall that OLS is the gold standard in this case, and the conservative behavior of the SAR allows us to obtain outputs similar to ML. In Figure \ref{fig:Ftest}, we display the p-value in this challenging case where the null hypothesis is not rejected. At the bottom of the figure, we provide the SAR test based on the detection threshold that mimics classical tests for the slope based on a significance level.
The optimistic results obtained by CV methods with a limited sample size, which could indicate their better ability for linearity detection, are indeed a consequence of poor control of FPs. To demonstrate this issue, we designed a putative task consisting of a regression problem with no correlation at all, i.e., a correlation level equal to 0. In this case, we analyze the real case with almost no repetitions ($R=2$), where the variability of errors in folds affects the evaluation of the F-statistic. In this scenario, we observe a non-smooth behavior of the F-statistic curves, but only ML CV lines cross below the significance level (e.g., $N=100$), unlike the SAR test or the F-test for the OLS estimation, which are always above the detection threshold, controlling the rate of FPs (see Figure \ref{fig:Ftest2}). Moreover, to illustrate this issue, we performed a power analysis in Figure \ref{fig:power1} with a wide variety of correlation levels, showing that CV methods are inflating the rate of FPs (detection rates above the OLS method for low correlation levels).

\subsection{Non-linearly transformed Gaussian data: large Cross-Validation residuals and False Positives}
\begin{figure}
\centering
\includegraphics[width=0.49\textwidth]{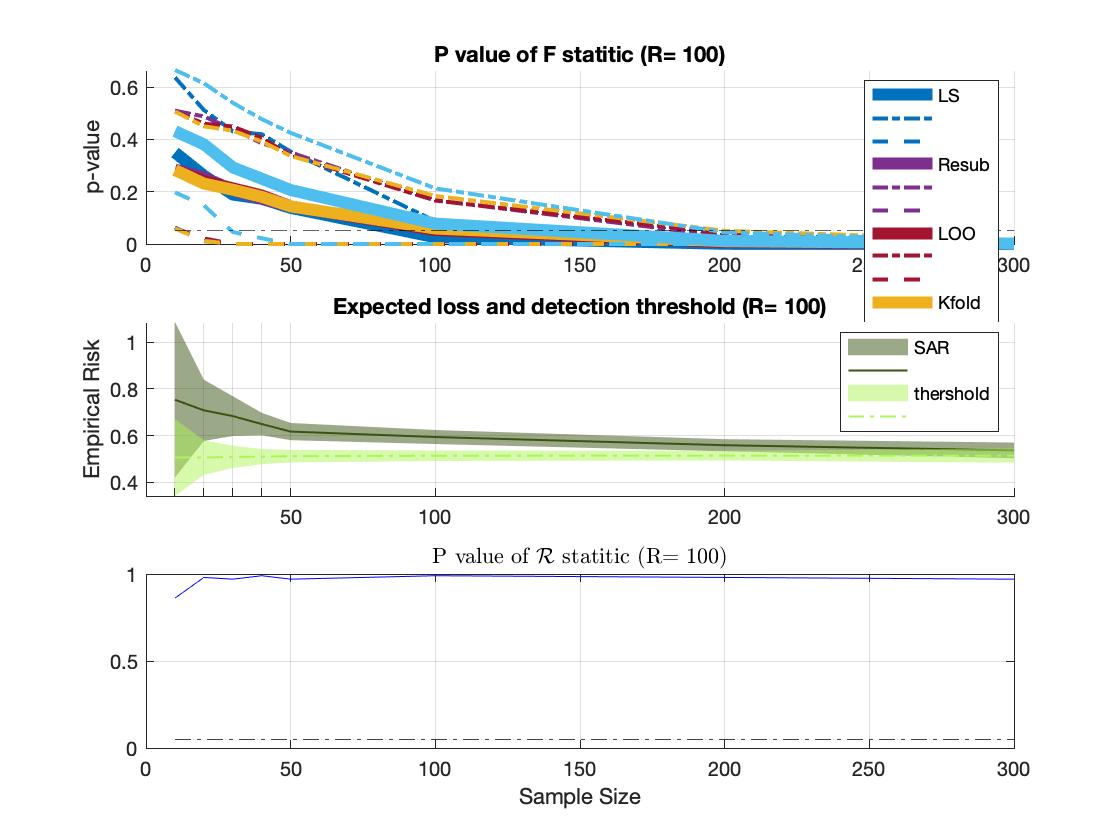}
\includegraphics[width=0.49\textwidth]{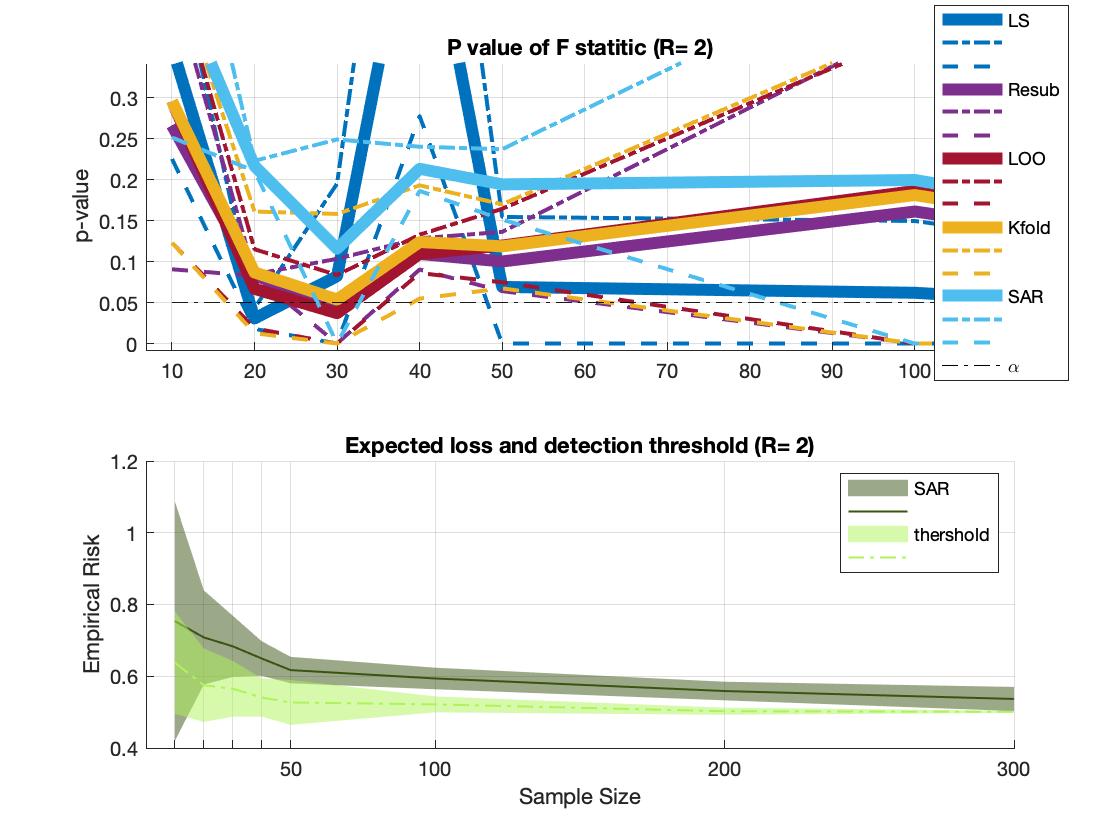}
\caption{The F-test for the slope $\upbeta_1$ in the case of correlation levels equal to $0.37$ with non-Gaussian data. We employed the residuals derived from each method and averaged the F-statistic and its variability in $R=100$ and $2$ repetitions for comparison purposes. } 
\label{fig:Ftest3}%
\end{figure}

In this case, the OLS method may not be the most efficient estimator, and inferences based on the normality assumption, such as the F-test, can lead to unreliable results. Here we find the true utility of the SAR test: to validate the significance of a prior regression approach. We analyzed two cases, as shown in Figure \ref{fig:datanG}, with correlation levels $\tau=0.63$ and $\tau=0.37$. Both exhibited similar p-values for the standard ML approaches and OLS. However, this is not always good news, as the Gaussianity assumption is not fulfilled. Particularly, for $R=100$ ideal realizations, from approximately $N=120$ and $\tau=0.37$, all methods rejected the null hypothesis, as shown in Figure \ref{fig:Ftest3}. One might be tempted to conclude that there is sufficient evidence for a linear relationship in the data. Unlike the SAR test, where the values of the expected loss converge to the detection threshold but always remain above it (p-value greater than the significance level). In realistic scenarios with $R=2$, that is, the typical hypothesis test using a classical approach, all methods provided a p-value above $0.05$ across a wide range of sample sizes, contravening the previous results. Thus, for example, results from a single laboratory with only two samples cannot be extrapolated to the rest of the realizations. In other words, depending on the repetition, we may have a significant result, or the null hypothesis cannot be rejected. As shown in the same figure, the SAR test, being conservative, is robust in this scenario compared to the F-test.

\begin{figure}
\centering
\includegraphics[width=0.49\textwidth]{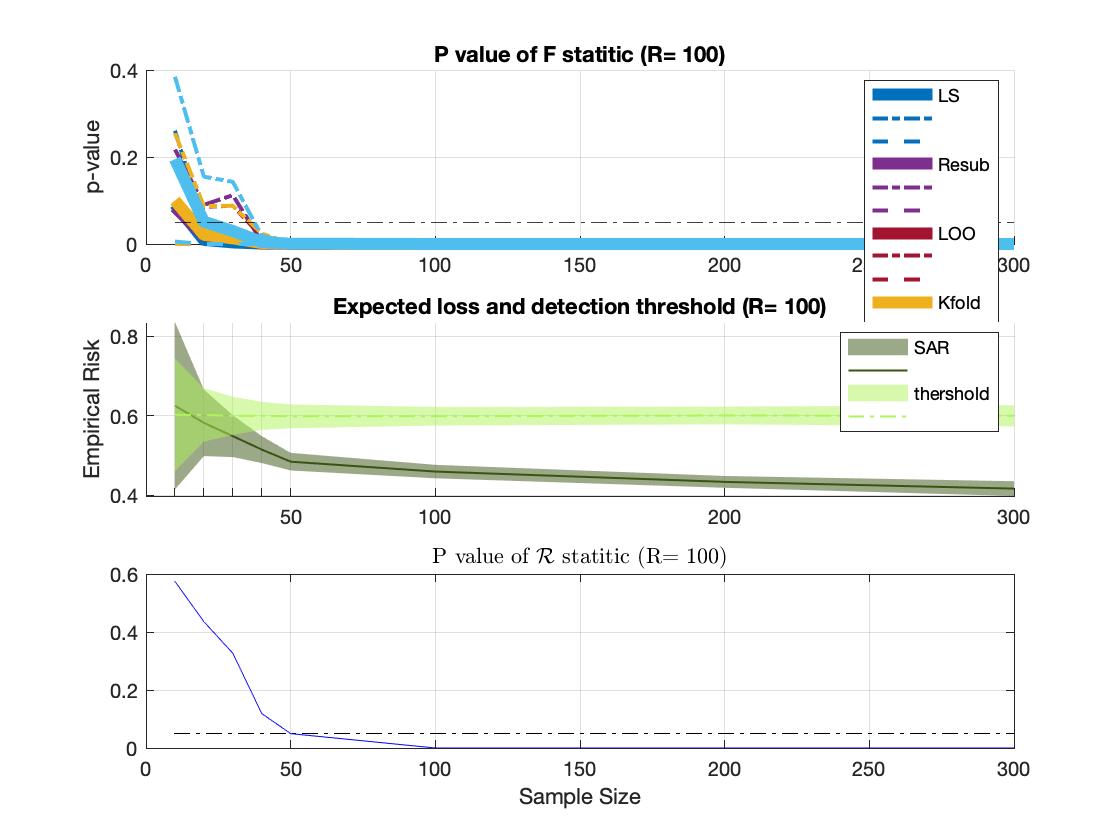}
\includegraphics[width=0.49\textwidth]{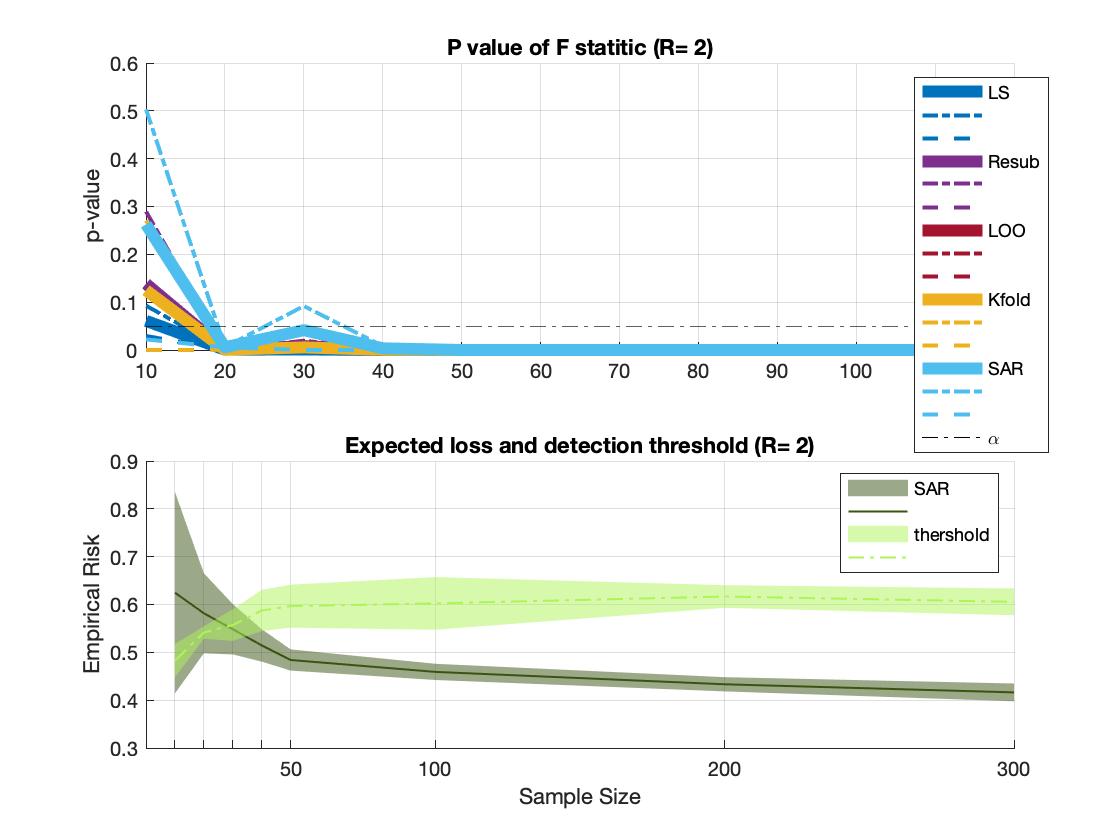}
\caption{The F-test for the slope $\upbeta_1$ in the case of correlation levels equal to $0.63$ with non-Gaussian data. We employed the residuals derived from each method and averaged the F-statistic and its variability in $R=100$ and $2$ repetitions for comparison purposes. } 
\label{fig:Ftest4}%
\end{figure}

\begin{figure}
\centering
\includegraphics[width=\textwidth]{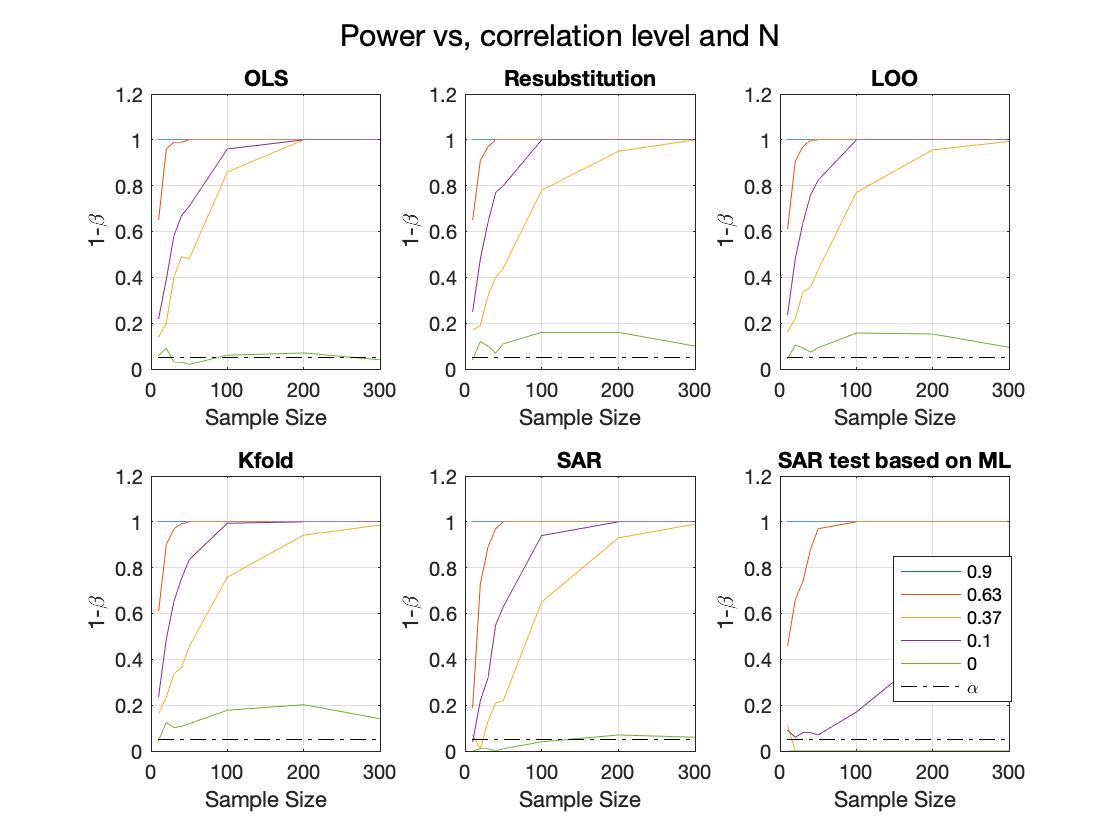}
\caption{Power analysis derived from the F-test in non-Gaussian data (except for the green solid line that represents uncorrelated data). The selected loss for ML methods is $\mathcal{L}_2$. Observe how the results obtained by the SAR features are by far the most similar to those obtained by the OLS for low correlation levels. However, for high levels, the performance of SAR is a trade-off between ML methods and OLS. The SAR test based on expected loss is more conservative than the other methods (correlation levels below $0.37$ are not considered significant based on the worst-case analysis with a probability of at least $\eta=0.5$).} 
\label{fig:power2}%
\end{figure}

Another example is shown in Figure \ref{fig:Ftest4} with $\tau=0.63$ (indicating significant linearity), where the SAR test validates the significance of the results from approximately $N=50$.  Note that the SAR test is robust against the number of repetitions in the bootstrap approach, since for $R=100$ and $R=2$ repetitions, we reject the null hypothesis at the same sample size, and above $N=50$, leading to good replication. Finally, we performed a power analysis, including a putative task with a correlation level equal to zero to assess how the methods control the rate of FPs \cite{Rosenblatt16}.  In this case, the rate of FPs should be approximately equal to the level of significance $\alpha$. Figure \ref{fig:power2} shows that the ML methods are inflating FPs (green solid line) above the level of significance, i.e., $(1-\beta)>\alpha$. Recall that the results of OLS under conditions of non-Gaussianity should be subject to scrutiny (except for the case $\tau=0$).

\subsection{Heteroscedascity data: superior detection using ML techniques}

\begin{figure}
\centering
\includegraphics[width=\textwidth]{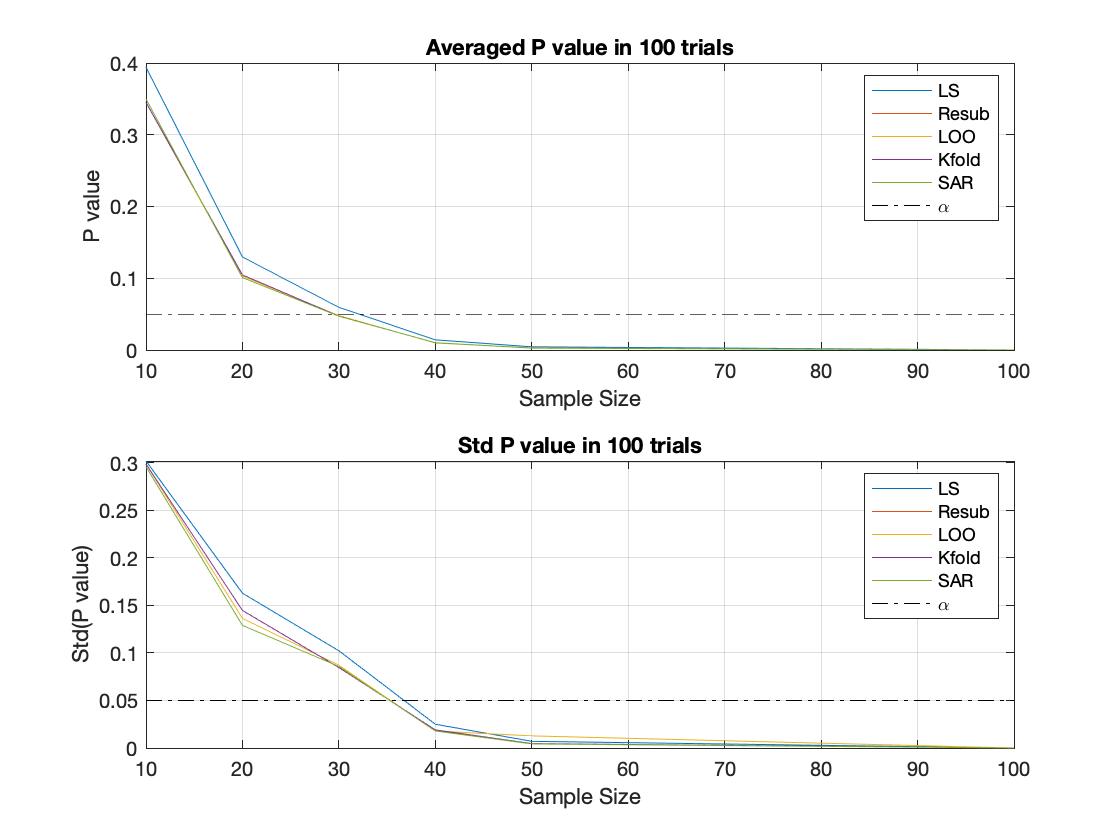}
\caption{The BP test on the residuals.}
\label{fig:BP2}%
\end{figure}

In this section, we tested the ability of the methods to detect heteroscedasticity using the BP test described in Section \ref{sec:matmet}. The problem analyzed here is clearly linear, but the distribution of residuals is not homogeneous (see Figure \ref{fig:BP1}). A simple Q-Q plot analysis of the data reveals the strong heteroscedasticity of the data,

We evaluated the BP test on the residuals obtained with all methods and compared the averaged and standard deviation (std) P-value with increasing sample size in $R=100$ repetitions. The BP test using the SAR residuals provides a faster detection of heteroscedasticity on average and a lower standard deviation of the P-value, as shown in Figure \ref{fig:BP2}.

\subsection{Multivariate Linear Regression: SAR features effectively detect linearity.}

\subsubsection{The Cancer Dataset}
\begin{figure*}
\centering
\includegraphics[width=0.49\textwidth]{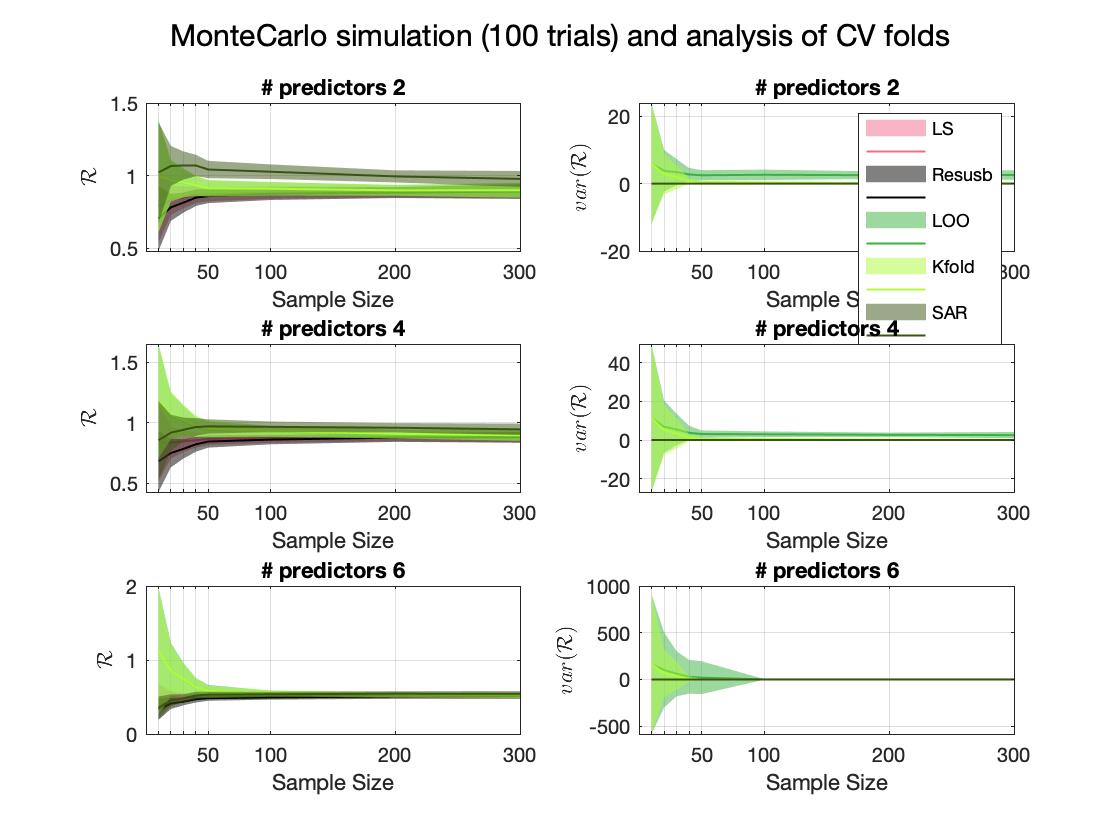}
\includegraphics[width=0.49\textwidth]{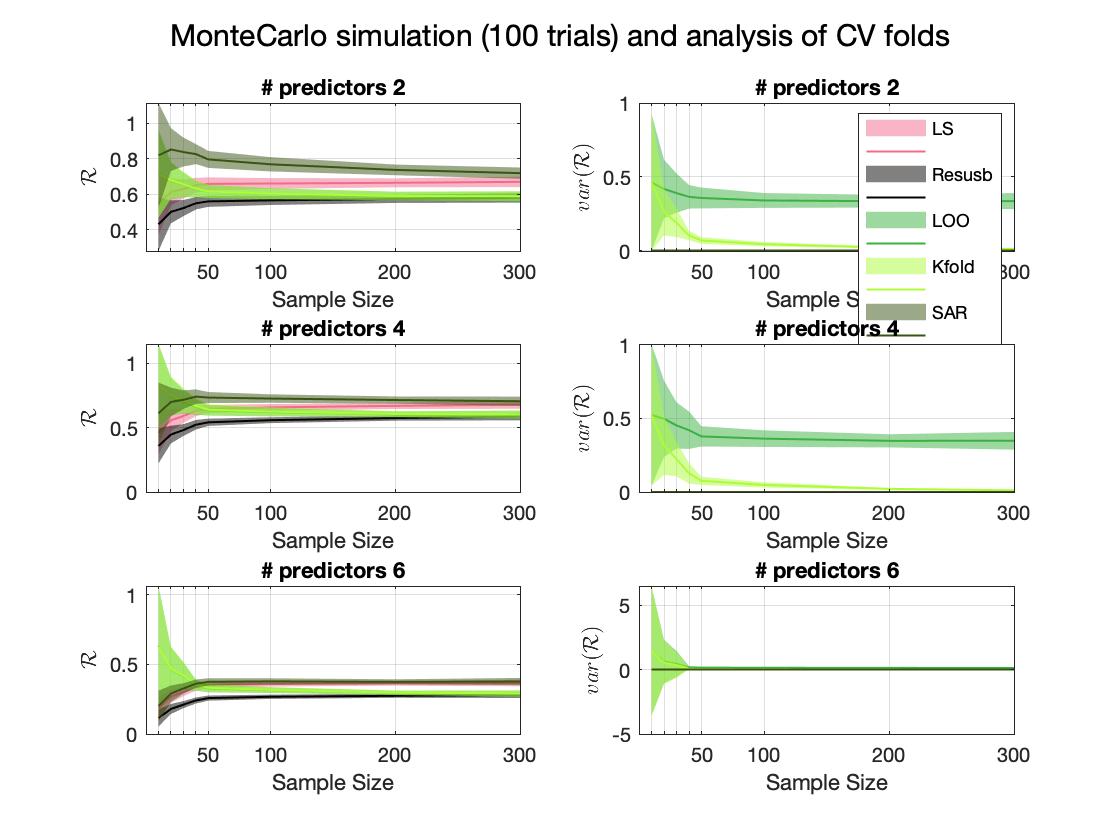}
\caption{Expected losses ($\mathcal{L}_2$ on the left and $\mathcal{L}_1$ on the right) and analysis of variance from folds. Again the variance using the $\mathcal{L}_2$ loss is greater than the  $\mathcal{L}_1$ loss for ML methods using CV.  However, with real data the presence of large outliers in folds results in inadmissible loss estimations.} 
\label{fig:datarealloss}%
\end{figure*}

We repeated the test for linearity and the power analysis with increasing dimensions, from one to six predictors, using the Cancer dataset. We observed two primary effects. First, the variance of CV methods is extremely large for small sample sizes ($N=10-50$), especially with the $\mathcal{L}_2$ loss, although it is significant for the rest of the simulations. This effect depends on the predictor added in the regression analysis. Notably, the last predictor that was added, 'Pop\_ estimate' (predictor 6), increases the variance of the estimators resulting in non-reliable estimates within training folds, mainly with small sample sizes (Figure \ref{fig:datarealloss}). The reason for this anomalous operation was previously shown in Figure \ref{fig:6c} in light blue. Here, we readily see in the SAR test (bottom of the figure) that evidence for a linear relationship in the data, under non-ideal conditions, is achieved with only 10 samples (the risk is always less than the threshold).

\begin{figure*}
\centering
\begin{subfigure}{0.49\textwidth}
    \includegraphics[width=\textwidth]{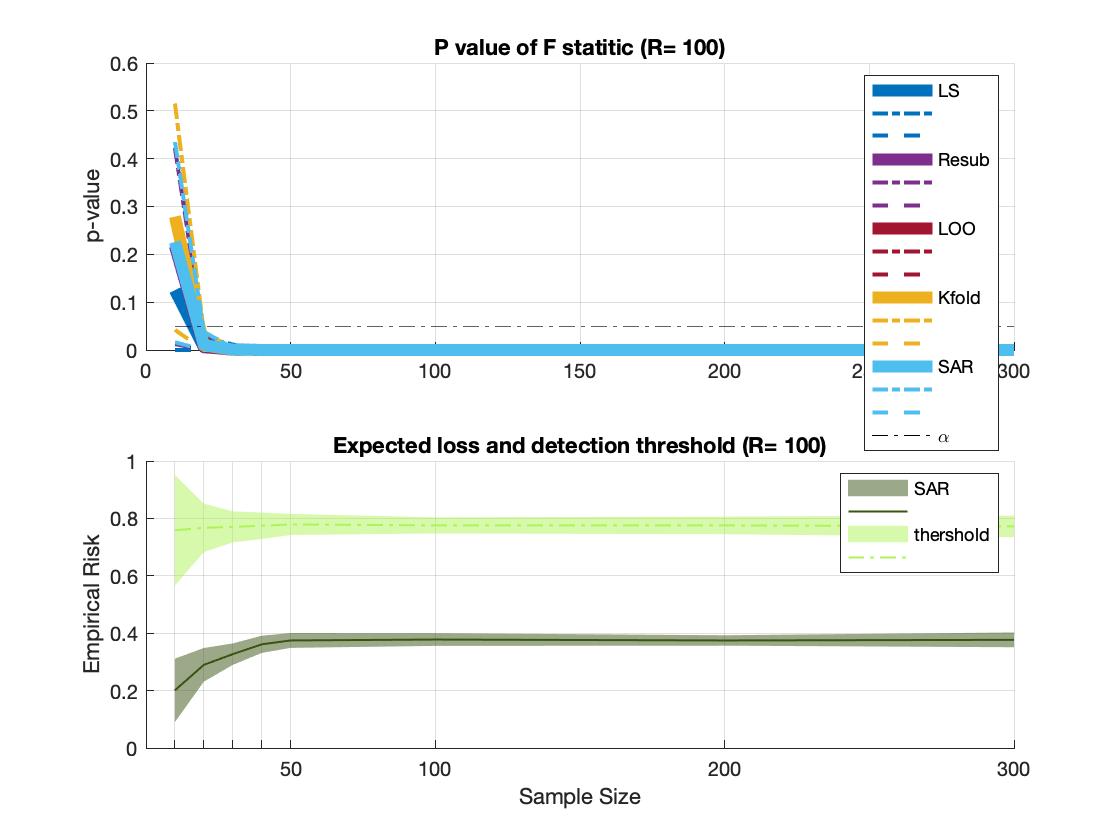}
    \caption{Testing the linear relationship in Cancer dataset with 6 predictors using $\mathcal{L}_1$.}
    \label{fig:7a}
\end{subfigure}
\hfill
\begin{subfigure}{0.49\textwidth}
    \includegraphics[width=\textwidth]{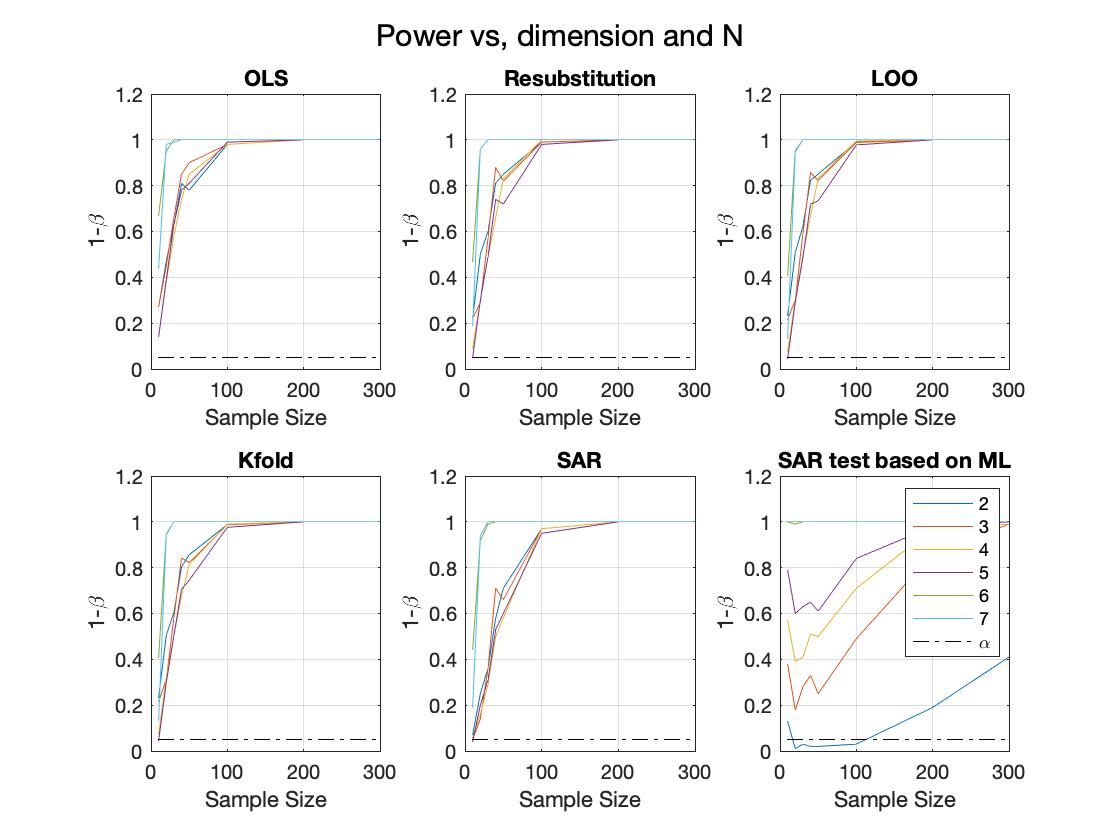}
    \caption{Power analysis versus the $\#$ predictors and sample size. Note that results are obtained with the most competitive loss for ML methods ($\mathcal{L}_1$).}
    \label{fig:7b}
\end{subfigure}
\hfill
\begin{subfigure}{0.49\textwidth}
    \includegraphics[width=\textwidth]{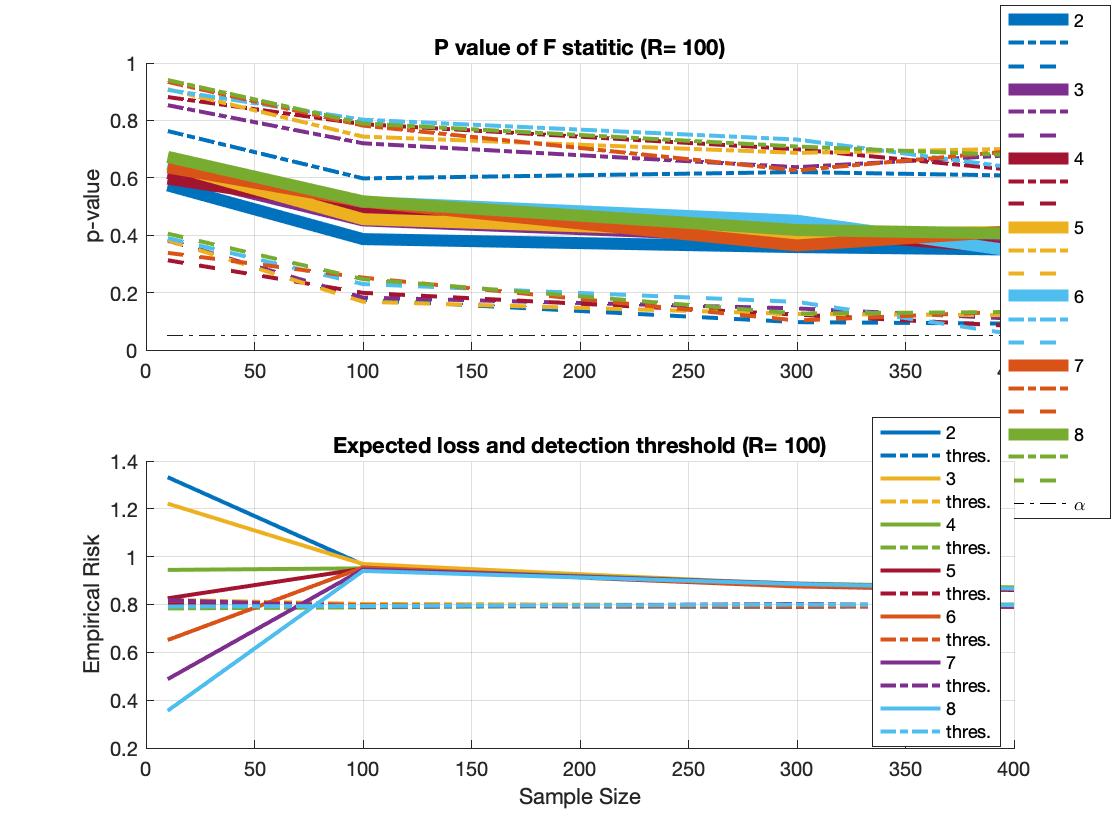}
    \caption{Test for linearity in the putative task of uncorrelated samples using PAC bound.}
    \label{fig:7c}
\end{subfigure}
\hfill
\begin{subfigure}{0.49\textwidth}
    \includegraphics[width=\textwidth]{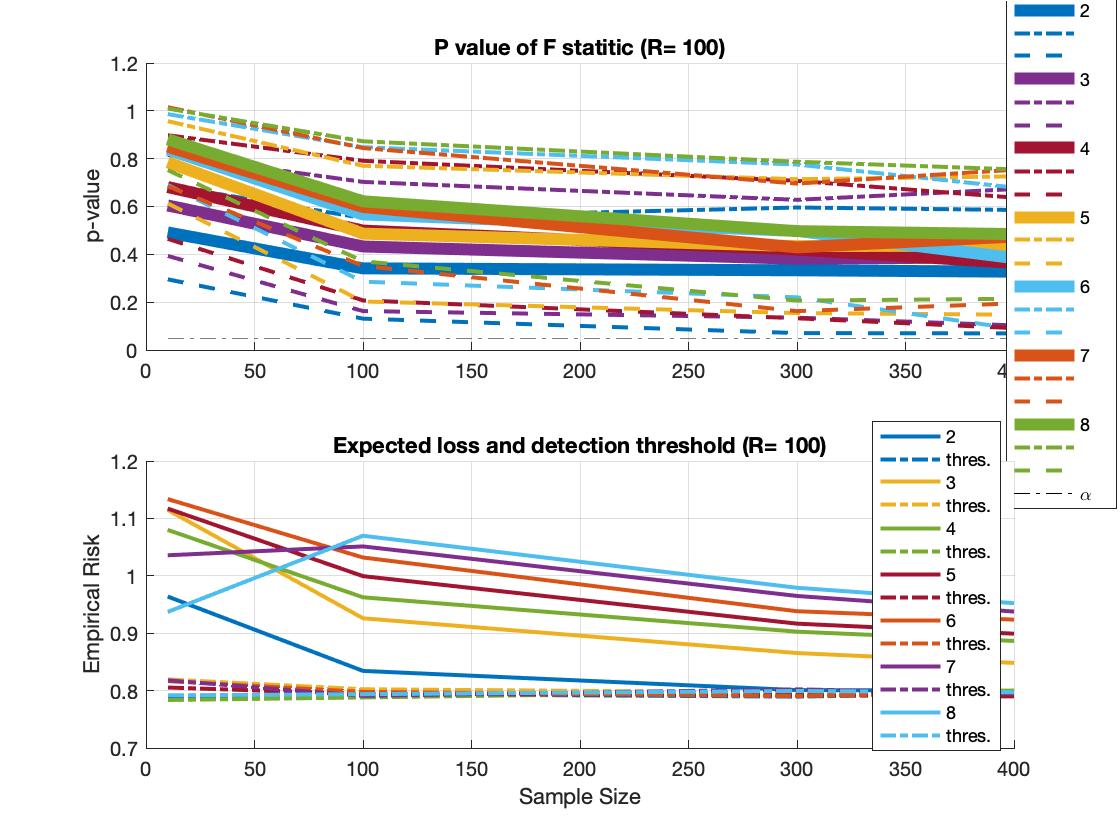}
    \caption{Test for linearity in the putative task of uncorrelated samples using the i.g.p.  bound.}
    \label{fig:7d}
\end{subfigure} 
\caption{Performance with increasing number of predictors in Cancer Dataset. }
\label{fig:cancer2}
\end{figure*}

The second effect worth commenting on is the detection ability of the SAR method when the number of predictors increases, as shown in Figure \ref{fig:7a}. Moreover, the SAR method converges to the OLS method with the two tested losses, unlike the CV-based ML methods. This suspicious feature found for the SAR test (and for any other ML method) can be explained by overfitting since it goes against its common behavior. To discard this possibility, we increased the dimensions or the number of predictors used in the simulation representing the putative task where there is no correlation. In this task, we additionally employed a general upper bound based on the assumption of samples in general position (i.g.p.), as proposed in \cite{Gorriz19}. The analysis for the two bounds is given in Figures \ref{fig:7c} and \ref{fig:7d}. As shown in this analysis, when we increase the number of dimensions (only in this section), the bound proposed in \cite{MacAllester2013} is overly optimistic compared to that proposed in \cite{Gorriz19} when $N<100$. In this case, PAC approaches should be reformulated, and analytical bounds such as those proposed in \cite{Vapnik82, Gorriz19} are valid solutions to formulate the SAR test as they effectively control the FP rate.

\subsubsection{The ADNI dataset}
\begin{figure*}
\centering
\begin{subfigure}{0.49\textwidth}
    \includegraphics[width=\textwidth]{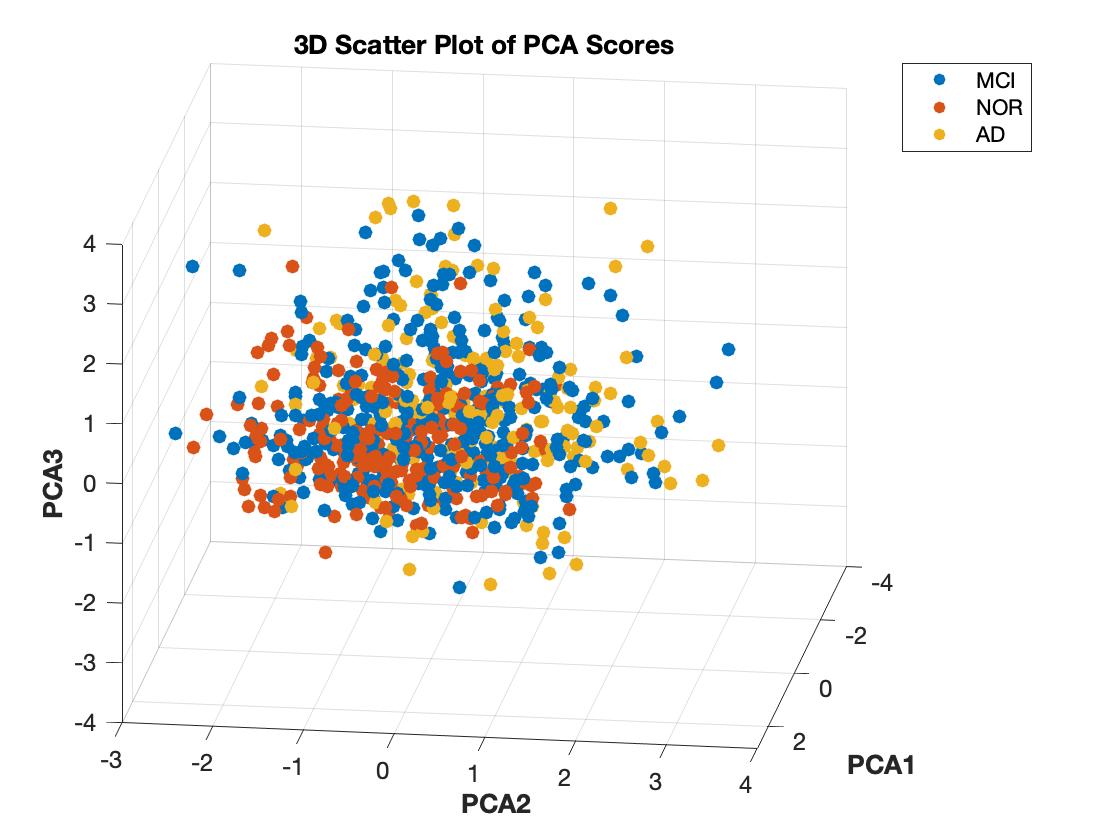}
    \caption{3D PCA clustering of the original ADNI dataset}
    \label{fig:7ADNIa0}
\end{subfigure}
\hfill
\begin{subfigure}{0.49\textwidth}
    \includegraphics[width=\textwidth]{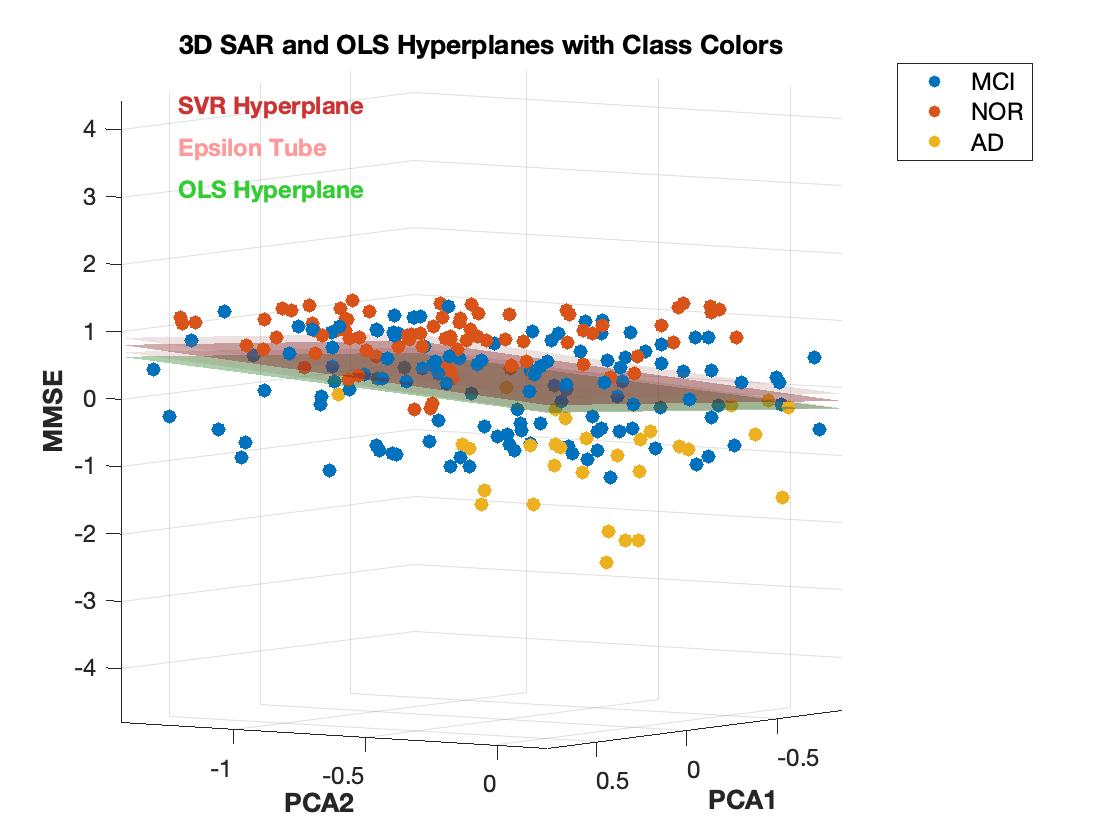}
    \caption{3D regression hyperplanes for MMSE vs predictors PCA 1-2.}
    \label{fig:7ADNIb0}
\end{subfigure}
\caption{PCA features in ADNI dataset (NOR,AD and MCI subjects).}
\label{fig:ADNI2}
\end{figure*}

The correlation between MMSE scores and MRI images is important because it provides valuable insights into the relationship between cognitive function and structural brain changes. We can clearly observe this effect in the set of PCA features extracted from the original dataset, prior to the regression analysis shown in Figure \ref{fig:7ADNIa0}. In the same figure, we present the regression fit using both OLS and SAR approaches (including the $\epsilon$-tube), where the observed MMSE scores and predictors are normalized using the z-score method. Note how the AD features are primarily located below the hyperplanes, while the NOR features are above them.

\begin{figure*}
\centering
\begin{subfigure}{0.49\textwidth}
    \includegraphics[width=\textwidth]{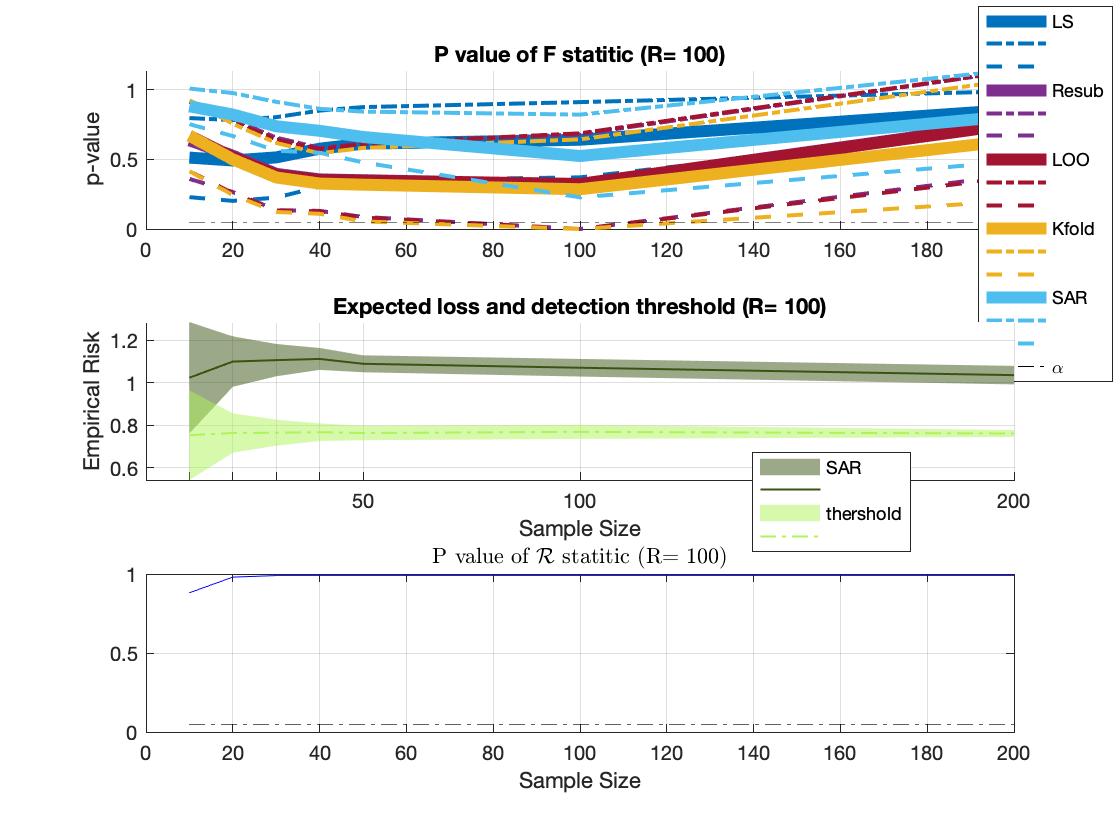}
    \caption{Testing the linear relationship in the NOR group ADNI dataset with 6 predictors using $\mathcal{L}_1$.}
    \label{fig:7ADNIa}
\end{subfigure}
\hfill
\begin{subfigure}{0.49\textwidth}
    \includegraphics[width=\textwidth]{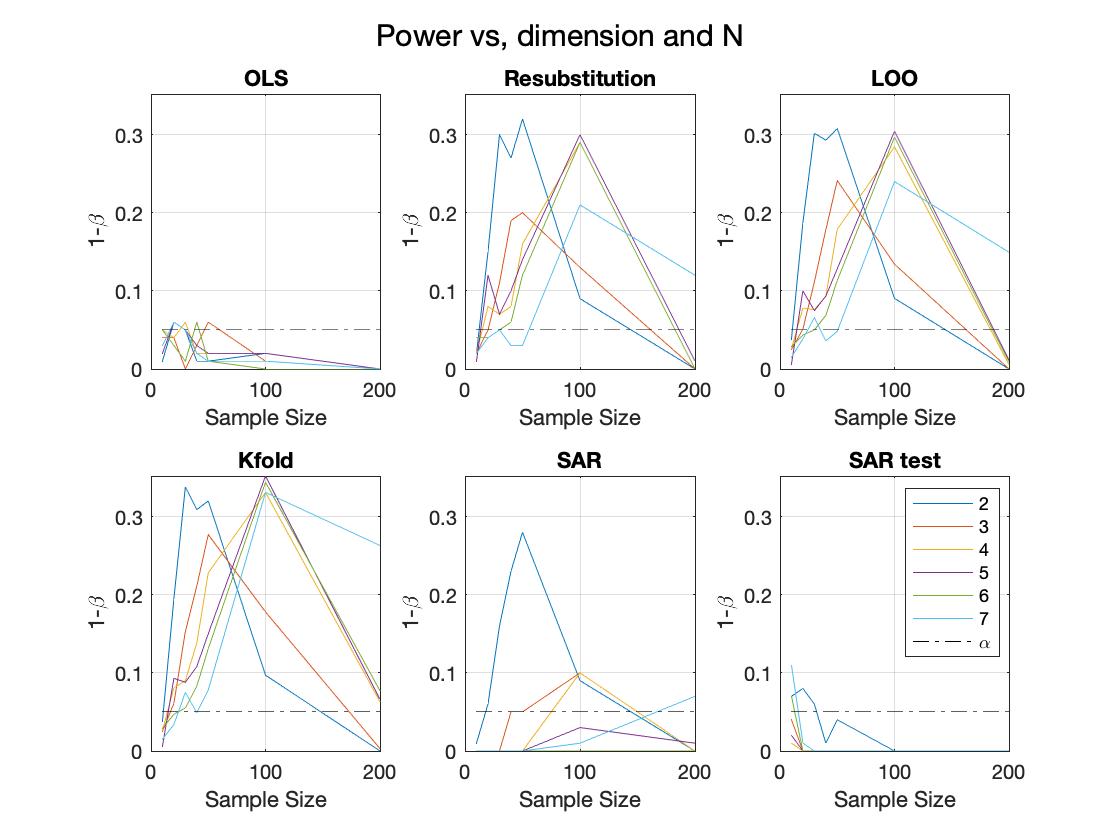}
    \caption{Power analysis versus the $\#$ predictors and sample size. Note that results are obtained with the most competitive loss for ML methods ($\mathcal{L}_1$).}
    \label{fig:7ADNIb}
\end{subfigure}
\hfill
\begin{subfigure}{0.49\textwidth}
    \includegraphics[width=\textwidth]{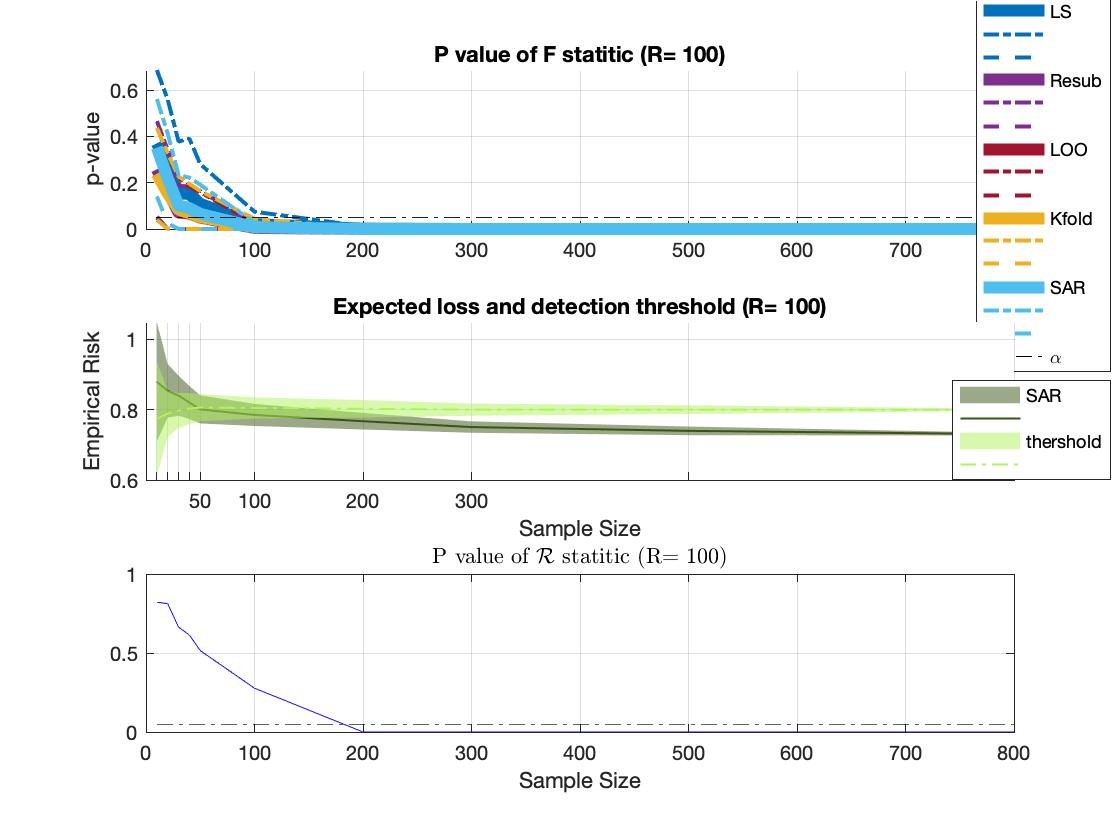}
    \caption{P-value analysis using $R$ repetitions in one dimension for the ADNI dataset}
    \label{fig:7ADNIc}
\end{subfigure}
\hfill
\begin{subfigure}{0.49\textwidth}
    \includegraphics[width=\textwidth]{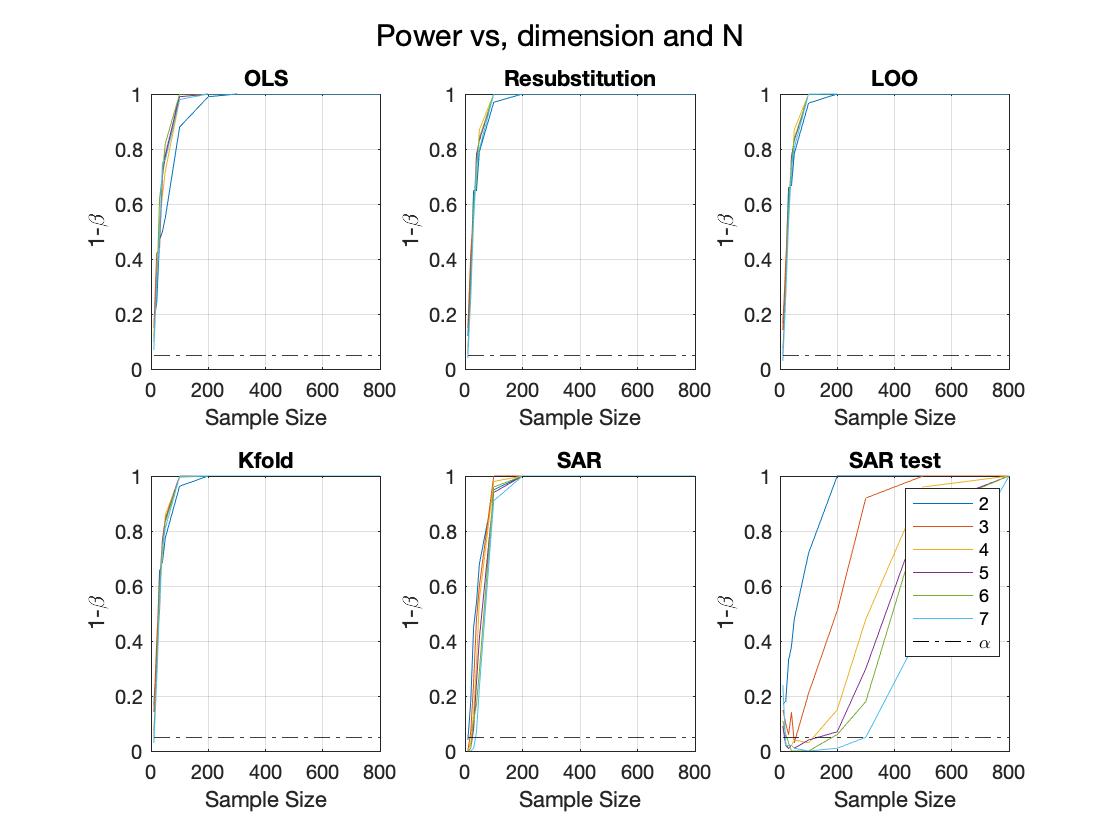}
    \caption{Power analysis with increasing dimensions using the ADNI dataset}
    \label{fig:7ADNId}
\end{subfigure} 
\caption{NOR and NOR-AD-MCI group analyses using the ADNI dataset.}
\label{fig:ADNI2}
\end{figure*}

We first analyzed this relationship in the control group (NOR) to assess the ``quality'' of the baseline group. The correlation between MMSE scores and MRI control images might not be as strong or consistent as in groups with MCI or AD. In healthy individuals, both MMSE scores and brain structures are likely to be within normal ranges, with less variability in cognitive function and brain structure. We detected a suspicious behaviour in the power of the F-test of the methods based on ML. They significantly correlated MMSE with the PCA features above the significance level (figure \ref{fig:7ADNIb}). This effect is essential for comparing against groups with neurodegenerative conditions like MCI or AD. If the control group shows unexpected correlations, it raises concerns about the validity of the baseline for future comparisons.

However, a p-value analysis over a set of $R$ repetitions showed that no significant correlation can be found between MMSE and the brain structure as expected (figure \ref{fig:7ADNIa}). Despite the ML models indicating a significant correlation, a p-value analysis conducted over multiple repetitions ($R$) contradicts this finding, showing that no significant correlation actually exists between MMSE and brain structure in the control group. This discrepancy highlights potential overfitting or spurious correlations within the ML methods when applied to group analyses in realistic experimental setups.

Finally, if we consider the three classes (AD, NOR, and MCI) and an increasing sample size $N=10$ to 800, we detect linearity with a few samples using the F-test, and above $N=200$ with the SAR test as shown in figure \ref{fig:7ADNIc}. Notice how the SAR test penalizes increasing dimensions, in contrast to classical residual-based tests for linearity (figure \ref{fig:7ADNId}).

\section{Discussion}

Initially, we recapitulated the conventional application of OLS in linear regression, emphasizing its widespread adoption due to its optimal statistical properties. The transition to ML introduces regularization highlighting the shift in focus towards minimizing both empirical risk and model complexity. Critically, we highlighted a common deficiency in machine learning approaches, particularly in regression, where rigorous significance analysis is often neglected in favor of permutation testing based on cross-validation measures, especially with small sample sizes. The proposed SAR method seeks to bridge this gap by formulating a statistical test grounded in statistical learning theory. This involves establishing an upper bound on the expected loss comparing it to the null hypothesis and rejecting it if the corrected risk is lower, signifying evidence of a linear relationship. 

The background of Support Vector regression was then explored presenting the $\epsilon$-insensitive loss function and the regularized risk functional. The discussion extends to exploring the flatness property and empirical loss, highlighting the differences in loss functions employed in ML models. Finally, the novel SAR method was introduced as a formal test for assessing the error estimation of ML algorithms, drawing parallels with classical statistical tests like the F-test. The section emphasized the significance of SAR in overcoming limitations in ML approaches and extended the application to regression analysis.

The results section presented a comprehensive examination of experimental outcomes focusing on the performance of the proposed SAR method in comparison to standard ML techniques across diverse scenarios. For Gaussian distributed data, where OLS serves as the gold standard, the SAR method exhibited conservatism converging to OLS with increasing sample sizes \cite{Noble20}. Testing with the F-test for the slope at a correlation level of $0.1$ revealed the optimistic behavior of ML methods, except for SAR, which struck a trade-off with OLS \cite{Phipson2010}. Analysis of non-Gaussian distributed data emphasized the utility of SAR in validating significance (reproducibility), especially in scenarios with limited experimental repetitions ($R=2$) \cite{NAS2019}. The detection of heteroscedasticity using the BP test demonstrated SAR's faster detection and lower standard deviation of the p-values. This thorough analysis, illustrated through various figures, underscores SAR's versatility and robustness in assessing error variability, validating significance, and detecting heteroscedasticity across different data scenarios (sample sizes). To sum up, from the set of experiments undertaken here, we demonstrated that the proposed SAR is effective for assessing the variability of error values obtained by CV in ML methods. Moreover, the evidence derived from these experiments suggests that SAR is a robust approach to functional regression.

As mentioned in the introduction, there is increasing concern about the use of K-fold CV as a baseline method for model selection. AI is proposing more complex algorithmic architectures that are validated with the easy to implement CV method, potentially leading to significant variability if sample sizes are limited or the data heterogeneity is large. This raises the question of whether the giant has feet of clay. When the data allows us to derive a stable inducer by splitting (making a guess from some parts of it), it can be a very robust method to derive models beyond the current realization. However, this assumption is not always met. By performing this procedure, we can expect that what we learn in the training folds is not necessarily useful for prediction with the remaining samples. This is analogous to the judgment of Solomon; we ask the two women to determine who the baby belongs to, and the information provided by each party should be useful. In case the two women reply the same, Solomon would have to sadly split the baby in two. In the biblical story, the reaction of each of the two women to the suggestion of splitting the baby was not the same, and Solomon gave the baby to the real mother. This is akin to what we have done with data using SLT, although we correct this resubstitution (split in two) by considering the information contained in the data (reaction from data). In our case, we would give the two parts to the more affected woman.

The theoretical findings are also applicable to real data in various fields of research. Indeed, the models devised in this paper, together with the simulation of realistic datasets, create suitable exemplars for characterizing performance in neuroimaging applications \cite{Gorgen18,Zhang14,Gorriz2021b,Gorriz2022,Wang07,Wang09}. A simple comparison in classification tasks between the simulated datasets and those of real situations reveals the similarity in the results obtained \cite{Gorriz19}. Nevertheless, the scatter plots and data distributions projected on the dimensions were clear examples demonstrating that the conditions to provide stable inducers \cite{Kohavi95} were not met. Therefore, exploration of alternative validation methods is a priority.

In summary, echoing practices from our previous work, we underscore the importance of emphasizing negative results to enhance scientific understanding\footnote{For further insights, refer to the column in Nature addressing this issue: https://www.nature.com/articles/d41586-019-02960-3}. While research papers often prioritize positive outcomes, we infrequently assess our algorithms in hypothetical task scenarios where no discernible effect is expected. This is evident in our experiments with uncorrelated data. Such analyses play a crucial role serving to approximate the null-distribution of the test-statistic in permutation studies, such as evaluating performance or accuracy in a classification task using ML techniques. For instance, in the permutation analysis of classification tasks, the performance derived from paired data and labels is juxtaposed with that obtained by randomly permuting group labels numerous times with the anticipated distribution centered around $50\%$. If the performance distribution exhibits non-symmetry around random chance and bias, conclusions drawn from the test data may be compromised. This suggests a disparity in data distribution between groups under the null hypothesis, violating the i.i.d. assumption. Consequently, the estimation of p-values might lead to inaccurate conclusions at the family-wise level \cite{Phipson2010}.

\section{Conclusions}

In this paper, we present a method for validating regression models in the field of machine learning and its applications. The method is related to the F-test of classical hypothesis testing for establishing significance in linear models. We demonstrate that standard ML methods for model validation tend to overinflate FPs, thus requiring these approaches to ensure good replication and extrapolation of results in limited sample sizes. When SAR features are incorporated into classical statistical frameworks, they provide a trade-off between OLS and ML paradigms with excellent control of FPs (around the level of significance).  Although these pipelines could face criticism when dealing with non-Gaussian data, we also conducted the formal test presented based on CIs. This test discarded regression problems with low correlation levels under the worst-case scenario and provided statistical significance for the rest of the cases. The use of this formal test is intended to be combined with classical hypothesis testing, allowing us to confirm the analysis with techniques that are not formally valid, or to inform the researcher that there is insufficient evidence from the perspective of SLT to establish such a linear relationship. This constitutes the formal definition of hypothesis testing: conclusions about the data can only be drawn when we reject the null hypothesis; otherwise, caution must be exercised with the findings.

%
%

\section*{Acknowledgments}
This research is part of the PID2022-137451OB-I00 and PID2022-137629OA-I00 projects, funded by the CIN/AEI/10.13039/501100011033 and by FSE+.

%
\appendix{}
\section*{A note on the theoretical losses presented in caption of figure \ref{fig:example1}}
In Figure \ref{fig:example1}, we stated that the theoretical expected loss on a uniform distribution of pairs ${y, \hat{y}}$ was $\mathcal{R}=\frac{a^2}{6b}$ for $\mathcal{L}_1$ and $\mathcal{R}=\frac{b^2+a^2}{3}$ for $\mathcal{L}_2$, where $a$ and $b$ represent the maximum values for $\hat{y}$ and $y$, respectively, when the number of predictors is $P=1$. In general, the expected loss is given by:
\begin{equation} 
\mathcal{R}=E[\mathcal{L}(f,x,y)]=\int \mathcal{L}(f,x,y) dP(x,y) 
\end{equation}
In this simple case with only two dimensions, assume that ${x, y}$ are a set of uncorrelated samples (uniformly) centered at the origin. Then, the difference $(\beta\cdot x-y)$ is distributed around zero and equally probable within a volume $\mathcal{V}$. Define $a$ and $b$ as the bounds for the uniformly distributed pairs $\{\hat{y}=\beta\cdot x,y\}$ contained in this volume. Then $\mathcal{V}=2a\times 2b$, and the expected loss can be computed as:
\begin{equation}\label{eq:Ru}
\mathcal{R}_u=\frac{1}{\mathcal{V}}\int \mathcal{L}(\hat{y}, y) d\hat{y}dy 
\end{equation}
\noindent where $\mathcal{V}=4ab$. If $\mathcal{L}(\hat{y}, y)=|y-\hat{y}|$, this expression simplifies to:
\begin{equation}\label{eq:Ru1}
\begin{array}{l}
  \mathcal{R}_u=\frac{1}{\mathcal{V}}\int_{-a}^{a}\int_{-b}^{\hat{y}} (\hat{y}-y) d\hat{y}dy   + \frac{1}{\mathcal{V}}\int_{-a}^{a}\int_{\hat{y}}^{b} (y-\hat{y}) d\hat{y}dy \\
  =\frac{1}{4ab}\left(\frac{1}{3}a^3 - b^2a\right)+\frac{1}{4ab}\left(\frac{1}{3}a^3 + b^2a\right) = \frac{1}{6}\frac{a^2}{b}    
\end{array} 
\end{equation}
If we instead choose $\mathcal{L}=(y-\hat{y})^2$, the loss equation \ref{eq:Ru} becomes:
\begin{equation}\label{eq:Ru2} 
\begin{array}{l}
\mathcal{R}_u=\frac{1}{\mathcal{V}}\int_{-a}^a\int_{-b}^{b}\mathcal{L}(\hat{y}, y) d\hat{y}dy \\= \frac{1}{4ab}\left(\frac{4}{3}b^3a + \frac{4}{3}ba^3\right) = \frac{1}{3}(b^2 + a^2) 
\end{array} 
\end{equation}
Note that we expect the solution to be flat ($\upbeta \sim 0$) with uncorrelated data. Consequently, as $a$ approaches zero, the solutions should converge to the mean value of the 1-D loss evaluated on the observed variable (half of the interval and one-third of the square, respectively). Any algorithmic deviation from this ideal solution yields the theoretical expected loss described in equations \ref{eq:Ru1} and \ref{eq:Ru2}. By examining Figure \ref{fig:example1}, we can evaluate how sampling and non-ideal flatness affect the convergence to these theoretical values.


\begin{thebibliography}{00}


\bibitem{Zelen62}
Marvin Zelen. Linear Estimation and Related Topics. in Survey of Numerical Analysis edited by John Todd, McGraw-Hill Book Co. Inc., New York, pp. 558-577. (1962)

\bibitem{Hilt1977}
Hilt, Donald E., et al. (1977). Ridge, a computer program for calculating ridge regression estimates. doi:10.5962/bhl.title.68934.

\bibitem{Tibshirani1996}
Tibshirani, Robert (1996). Regression Shrinkage and Selection via the lasso. Journal of the Royal Statistical Society. Series B (methodological). 58 (1). Wiley: 267–88. 

\bibitem{Burges98}
C.J.C Burges. A tutorial on support vector machines for pattern recognition
Data Mining and Knowledge Discovery, 2 (2) (1998), pp. 121-167

\bibitem{Grohs22}
P Grohs, et al. Mathematical Aspects of Deep Learning. Cambridge University Press. ISBN 9781009025096. https://doi.org/10.1017/9781009025096.

\bibitem{Snee77}
Ronald D. Snee. Validation of Regression Models: Methods and Examples.  Technometrics. Vol. 19, No. 4 (Nov., 1977), pp. 415-428 

\bibitem{Kohavi95}
R. Kohavi.  A study of cross-validation and bootstrap for accuracy estimation and model selection. International Joint Conference on Artificial Intelligence (IJCAI), pp 1--7, 1995.

\bibitem{Kleijnen96}
J.P.C. Kleijnen et al.  Validation of trace-driven simulation models: regression analysis revisited.  Proceedings Winter Simulation Conference. . 352-359. 0-7803-3383-7.  Dec 1996.

\bibitem{Miller91}
Michael E. Miller et al. Validation techniques for logistic regression models. Statistics in Medicine. Volume 10, Issue 8  1213-1226. August 1991.

\bibitem{Oredein11}
A.I Oredein et. al. On Validating Regression Models with Bootstraps and Data Splitting Techniques. Global Journal of Science Frontier Research Volume 11 Issue 6 Version 1.0 September 2011

\bibitem{Moore2003}
Moore, D. S., et al (2003): Bootstrap Methods and Permutation Tests. In The Practice of Business Statistics Companion, chap 18. W. H. Freeman; First Edition ISBN 978-0716757269. 

\bibitem{Lecun15}
Y. LeCun et al. Deep learning. Nature 521, 436–444 (2015). 

\bibitem{Gorriz2020}
J.M.Gorriz,  et al.  Artificial intelligence within the interplay between natural and artificial computation: Advances in data science, trends and applications. Neurocomputing Volume 410, 14 October 237-270 2020.

\bibitem{Varoquaux18}
G. Varoquaux. Cross-validation failure: Small sample sizes lead to large error bars. NeuroImage 180 (2018) 68-77.

\bibitem{Eklund16}
A.Eklund, et al. Cluster failure: Inflated false positives for fMRI. Proceedings of the National Academy of Sciences Jul 2016, 113 (28) 7900-7905.

\bibitem{Jollans2019}
Jollans L,et al. Quantifying performance of machine learning methods for neuroimaging data. Neuroimage. 2019 Oct 1;199:351-365. 

\bibitem{Gorriz18}
J.M.Górriz, et al. A Machine Learning Approach to Reveal the NeuroPhenotypes of Autisms. International journal of neural systems, 1850058. 2019.

\bibitem{Braga04}
Braga-Neto UM, Dougherty ER. Is cross-validation valid for small-sample microarray classification? Bioinformatics. 2004 Feb 12;20(3):374-80. doi: 10.1093/bioinformatics/btg419. PMID: 14960464

\bibitem{vandermaaten08}
L.van der Maaten et al. Visualizing Data using t-SNE. Journal of Machine Learning Research 2008 vol 9, num 86, 2579--2605.

\bibitem{Jimenez22}
C. Jimenez-Mesa et al. A non-parametric statistical inference framework for Deep Learning in current neuroimaging. Information Fusion Volume 91, March 2023, Pages 598-611.

\bibitem{Gorriz2021}
J.M.Gorriz, et al.  Statistical Agnostic Mapping: A framework in neuroimaging based on concentration inequalities.  Information Fusion Volume 66, February 2021, Pages 198-212

\bibitem{Viering23}
Tom Viering et al.  The Shape of Learning Curves: A Review. IEEE Transactions on Pattern Analysis and Machine Intelligence
June 2023, pp. 7799-7819, vol. 45 DOI Bookmark: 10.1109/TPAMI.2022.3220744

\bibitem{Vapnik95}
V. Vapnik. The nature of statistical learning theory Springer-Verlag New York, Inc., (1995)

\bibitem{Scholkopf95}
Bernhard Scholkopf et. al. Learning with Kernels: Support Vector Machines, Regularization, Optimization, and Beyond.  MIT Press
ISBN:978-0-262-19475-4. December 2001

\bibitem{Huber64}
Huber, P. J. (1964).  Robust estimation of a location parameter. Ann. Math. Statist., 35, 73–101

\bibitem{Chatterjee06}
Chatterjee, S. and Hadi, A.S.  Regression Analysis by Example. 4th Edition, John Wiley \& Sons, Hoboken.
https://doi.org/10.1002/0470055464. (2006) 

\bibitem{Bullmore99}
E T Bullmore et al.  Global, voxel, and cluster tests, by theory and permutation, for a difference between two groups of structural MR images of the brain IEEE Trans Med Imaging (1999) Jan;18(1):32-42.  

\bibitem{Reiss15}
P.T. Reiss,  et al. Cross-validation and hypothesis testing in neuroimaging: an irenic comment on the exchange between Friston and Lindquist et al. Neuroimage. 2015 August 1; 116: 248-254

\bibitem{Gorriz23}
Juan M Gorriz et al. Is K-fold cross validation the best model selection method for Machine Learning? arXiv:2401.16407

\bibitem{Vapnik82}
V. Vapnik. Estimation dependencies based on Empirical Data. Springer-Verlach. 1982 ISBN 0-387-90733-5

\bibitem{Gorriz19}
JM Górriz, et al. . On the computation of distribution-free performance bounds: Application to small sample sizes in neuroimaging
Pattern Recognition 93, 1-13 (2019)

\bibitem{Haussler92}
D. Haussler.  Decision theoretic generalizations of the PAC model for neural net and other learning applications. Information and Computation Volume 100, Issue 1, September 1992, Pages 78-150

\bibitem{Boucheron13} 
S. Boucheron et al.  Concentration Inequalities: A Nonasymptotic Theory of Independence ISBN: 9780199535255 Oxford University Press

\bibitem{MacAllester2013}
D. McAllester,  A PAC-Bayesian tutorial with a dropout bound.  arXiv 10.48550/ARXIV.1307.2118,2013

\bibitem{Breusch79}
T. S. Breusch et al. A Simple Test for heteroscedasticity and random coefficient variation. Econometrica , Sep., 1979, Vol. 47, No. 5 (Sep., 1979), pp. 1287-1294

\bibitem{Koenker81}
R. Koenker. A note on studentizing a  test for heteroscedascity. Journal of Econometrics 17. 107-l 12. North-Holland Publishing Company (1981).

\bibitem{Friston13}
K.J. Friston. Sample size and the fallacies of classical inference.  NeuroImage 81 (2013) 503–504.

\bibitem{Soft19}
Addinsoft, 2019. XLSTAT statistical and data analysis solution, Long Island, NY, USA. <https://www.xlstat.com>

\bibitem{Rosenblatt16} 
J.D. Rosenblatt, et al. Better-than-chance classification for signal detection. Biostatistics (2016).

\bibitem{NAS2019}
National Academies of Sciences, Engineering, and Medicine. (2019). Reproducibility and Replicability in Science. Washington, DC: The National Academies Press. https://doi.org/10.17226/25303.

\bibitem{Noble20}
S. Noble, et al. Cluster failure or power failure? Evaluating sensitivity in cluster-level inference. NeuroImage, 209, 116468,2020.

\bibitem{Phipson2010}
B. Phipson et al. Permutation P-values Should Never Be Zero: Calculating Exact P-values When Permutations Are Randomly Drawn. Statistical Applications in Genetics and Molecular Biology: Vol. 9: Iss. 1, Article 39. (2010)

\bibitem{Gorgen18}
Gorgen, K., et al. The same analysis approach: Practical protection against the pitfalls of novel neuroimaging analysis methods. NeuroImage, 180, 19-30. 2018.

\bibitem{Zhang14}
Y. Zhang et al. Multivariate lesion-symptom mapping using support vector regression. Hum Brain Mapp. 2014 Dec;35(12):5861-76. 

\bibitem{Gorriz2021b}
JM Gorriz,  et al. A connection between pattern classification by machine learning and statistical inference with the General Linear Model. IEEE Journal of Biomedical and Health Informatics 2021.

\bibitem{Gorriz2022}
JM Gorriz,  et al. A hypothesis-driven method based on machine learning for neuroimaging data analysis. Neurocomputing Volume 510, 21 October 2022, Pages 159-171

\bibitem{Wang07}
Z Wang,  et al.  Support vector machine learning-based fMRI data group analysis. NeuroImage 36 (4), 1139-1151. 2007

\bibitem{Wang09}
Z Wang.  A hybrid SVM–GLM approach for fMRI data analysis. Neuroimage 46 (3), 608-615. 2009.



\end{thebibliography}
\end{document}